%% file: main_preprint.tex
\documentclass[letterpaper]{article}
\usepackage{aaai2026}
\nocopyright
\usepackage{times}
\usepackage{helvet}
\usepackage{courier}
\usepackage[hyphens]{url}
\usepackage{graphicx}
\usepackage{natbib}
\usepackage{caption}
\usepackage{microtype}
\usepackage{booktabs} 
\usepackage{xcolor}

\usepackage{amsmath}
\usepackage{amssymb}
\usepackage{mathtools}
\usepackage{amsthm}

\usepackage{amsfonts}
\usepackage{dsfont}
\usepackage{multirow}
\usepackage{colortbl}
\usepackage{tabularx}
\usepackage{makecell}

\usepackage[capitalize,noabbrev]{cleveref}

\theoremstyle{plain}
\newtheorem{theorem}{Theorem}[section]
\newtheorem{proposition}[theorem]{Proposition}
\newtheorem{lemma}[theorem]{Lemma}
\newtheorem{corollary}[theorem]{Corollary}
\theoremstyle{definition}
\newtheorem{definition}[theorem]{Definition}
\newtheorem{assumption}[theorem]{Assumption}
\theoremstyle{plain}
\newtheorem{remark}[theorem]{Remark}

\title{Decision Potential Surface: A Theoretical and Practical
Approximation of\\ Large Language Model Decision Boundary}

\author{
Zi Liang\textsuperscript{\rm 1},
Zhiyao Wu\textsuperscript{\rm 2},
Haoyang Shang\textsuperscript{\rm 3},
Yulin Jin\textsuperscript{\rm 1},
Qingqing Ye\textsuperscript{\rm 1},
Huadi Zheng\textsuperscript{\rm 4},\\
Peizhao Hu\textsuperscript{\rm 5},
Haibo Hu\textsuperscript{\rm 1,6}\thanks{Corresponding author.}
}
\affiliations{
\textsuperscript{\rm 1}The Hong Kong Polytechnic University\\
\textsuperscript{\rm 2}University of Macau\\
\textsuperscript{\rm 3}Shanghai Jiaotong University\\
\textsuperscript{\rm 4}Huawei\\
\textsuperscript{\rm 5}Rochester Institute of Technology\\
\textsuperscript{\rm 6}PolyU Research Centre for Privacy and Security Technologies in Future Smart Systems\\
\texttt{\{zi1415926.liang,yulin.jin\}@connect.polyu.hk, bc31072@um.edu.mo}\\
\texttt{assassin808@sjtu.edu.cn, \{qqing.ye,haibo.hu\}@polyu.edu.hk}\\
\texttt{zhenghuadi@huawei.com, hxpvcs@rit.edu}
}

\begin{document}

\maketitle

\begin{abstract}
Decision boundary, the subspace of inputs where a machine
learning model assigns equal classification probabilities to
two classes, is pivotal in revealing core model properties
and interpreting behaviors.
While analyzing the decision boundary of large language models (LLMs)  
has attracted increasing attention recently, constructing it for
mainstream LLMs remains computationally infeasible due to the enormous
sequence-level output spaces and the autoregressive nature of LLMs.
To address this issue, in this paper we propose \emph{Decision Potential
Surface (DPS)}, a new notion for analyzing the properties of LLM
decisions. DPS is derived from the confidence in distinguishing
different classes for each input, which naturally
captures the \emph{potential} of the decision boundary.
We prove that the zero-height isohypse in DPS is equivalent to the
decision boundary of an
LLM, with enclosed regions representing decision regions.
By leveraging DPS, for the first time in the literature, we propose a practical
decision boundary approximation algorithm, namely $K$-DPS, which only
requires only $K$ finite sequence samples to approximate an LLM's
decision boundary with negligible error. We theoretically derive the
upper bounds for the absolute error, expected error, and the error
concentration between $K$-DPS and the ideal DPS, demonstrating that
such errors can be traded off against sampling times.
\end{abstract}


\input{intro2}
\input{related}

\input{method}

\input{method2}
\input{exper}

\section{Conclusion}
We propose DPS, a decision-boundary formulation for LLM sequence
generation, and $K$-DPS, a finite-sampling approximation that makes
boundary construction computationally feasible. We prove that the
zero-height isohypse of DPS recovers the decision boundary and derive
absolute, expected, concentration, and local candidate-set error
bounds. Experiments further show that $K$-DPS converges with practical
sampling budgets and reveals interpretable boundary changes under
alignment and unlearning.


\bibliography{refs}

\clearpage

\appendix
\input{appendix}

\end{document}

%% file: intro2.tex
\section{Introduction}\label{sec:intro}

With the rapid advancement and remarkable success of large language
models (LLMs), understanding their underlying mechanisms and behaviors
has become increasingly
critical~\citep{circuit1,circuit2,circuit3,circuit-overall,math-ana,physics-llm-3,lora-robustness,prompt-analysis}. A
key approach to demystifying the ``black box'' of state-of-the-art AI
models involves analyzing the \emph{decision
boundary}~\citep{rosenblatt1958perceptron}, a fundamental concept for
elucidating the characteristics of machine learning (ML) models. For LLMs,
decision boundaries provide valuable insights into critical phenomena,
including reasoning~\citep{db-hallucination}, in-context
learning~\citep{db-icl}, hallucination~\citep{db-memorization},
memorization~\citep{db-knowledge}, and so on.

As a foundational concept in machine learning, the decision
boundary represents a subspace of inputs where a model assigns equal
probability to two distinct classification
outcomes~\citep{rosenblatt1958perceptron}. Extensive
theoretical and empirical
studies~\citep{lee1997decision,turner1996analysis,goodfellow2015explaining,madry2018towards,gu2017badnets}
have demonstrated that the properties of decision boundaries reveal
critical attributes of machine learning models, including performance,
robustness, and generalization. Consequently, 
constructing and leveraging decision boundaries for LLMs
becomes a powerful and promising approach to enhancing
{\it almost all downstream analyses} of their behavior and
capabilities.

Unfortunately, analyzing decision boundaries of LLMs incurs significantly greater
complexity than for deep neural
networks (DNNs)~\citep{db1,db2,db3,lee1997decision,db5,db6}. Unlike 
classification tasks with a limited number of classes~\citep{lee1997decision,turner1996analysis,goodfellow2015explaining,madry2018towards,gu2017badnets},
LLMs predict a single token from an expansive vocabulary, often
exceeding 100,000 tokens. Moreover, their autoregressive
nature~\citep{lm,gpt} requires iterative token predictions to generate
complete sequences, which further compounds the complexity of modeling
decision boundaries. For instance, a Qwen-3 model (with 8 billion
parameters)~\citep{qwen3} supports sequences up to 32,768 tokens with
a vocabulary of 151,936, resulting in approximately $10^{169,790}$
decision regions! Such an enormous scale renders trivial attempts on
decision-boundary-based analysis and
visualization computationally infeasible.
Prior studies~\citep{db-icl,db-hallucination,db-memorization,db-knowledge},
despite their valuable contributions to their specific motivating tasks, unfortunately
sidestep this critical challenge.
They either simplify the
problem to toy scenarios, such as binary
classification~\citep{db-icl,db-memorization}, or use the decision
boundary concept metaphorically without constructing
it~\citep{db-hallucination,db-knowledge}. Consequently, the haunting 
questions remain unanswered: \textbf{What constitutes an LLM's decision
boundary, and is there a universal yet efficient algorithm to
construct it?}

To address these questions, we propose a principled strategy for
modeling the decision boundaries of LLMs, which yields
theoretical guarantees, computational tractability, and
interpretability simultaneously.
Inspired by the existing decision boundaries for multi-class
classification, we treat generative
language models as a composite multi-class classification
task. As trivial solutions cannot model the complex decision boundaries for
such tasks, 
we introduce a novel concept, namely \emph{Decision
  Potential Surface (DPS)}, to facilitate decision boundary
analysis. It is a landscape in which every point encodes the
\emph{competition potential} among candidate outputs, quantified by
a \emph{decision potential function (DPF)}.
We theoretically demonstrate that the zero-height
\emph{isohypse} of the DPS corresponds to the decision boundary, with the
enclosed regions representing decision regions. 

By examining the definition of DPS, we surprisingly discover that
enumerating the entire output space is unnecessary for computing the
DPF. Instead, sufficient sampling already captures the ``competition
potential''. We therefore approximate the LLM's decision boundary
with only $K$ finite ($K\ll$ realistic classification count) sequence
sampling, yielding \emph{$K$-DPS} and keeping the theoretical error within a
provably small bound.
We establish the error bound, expected error
bound, and error concentration between the ideal DPS and $K$-DPS,
demonstrating that $K$-DPS offers a favorable trade-off between
approximation accuracy and computational cost. Finally, we
conduct extensive experiments on open-source LLMs to evaluate the empirical
performance of our method.

To our best knowledge, this is the first study on
constructing decision boundaries for LLMs. Moreover, our proposed
\emph{decision potential surface (DPS)} framework is the first to
provide a practical approximation of decision boundaries with
theoretical guarantees. Our contributions are as follows:

\begin{itemize}
\item We introduce the concepts of the \emph{Decision Potential
    Function (DPF)} and \emph{Decision Potential Surface (DPS)}. We
  prove that the \emph{isohypses} of the decision potential surface represent the marginal decision
  boundaries of LLMs, with the zero-height isohypse equivalent
  to the decision boundary.
\item We propose $K$-DPS, an efficient and bounded approximation of the ideal DPS that
  requires only finite sampling for each input. We theoretically and
  empirically establish the error bounds of this approximation
  relative to the ideal DPS and quantify the trade-off between
  approximation error and sampling size.
\item Leveraging $K$-DPS, we present several insightful case studies
  that demonstrate how LLMs can be analyzed through the lens of their
  decision boundary properties.
\end{itemize}


%% file: related.tex
\section{Related Work}
Decision-boundary analysis has long been used to understand model
behavior, from linear classifiers~\citep{rosenblatt1958perceptron} to
feedforward networks~\citep{lee1997decision,turner1996analysis} and
modern deep models. Prior work links boundary geometry to robustness,
capacity, and failure modes: adversarial examples exploit locally
linear boundaries~\citep{goodfellow2015explaining}, adversarial
training smooths them~\citep{madry2018towards}, and backdoor attacks
can be understood as hidden boundary shifts~\citep{gu2017badnets}.
Subsequent studies further develop boundary extraction, visualization,
and quantitative metrics for trained networks~\citep{db1,db2,lee1997decision,db5,db6}.
For LLMs, boundary-oriented analysis remains less developed.
Existing studies investigate decision behavior in restricted settings,
including in-context learning~\citep{db-icl}, confidence and
counterfactual reliability~\citep{db-memorization}, and
boundary-aware reasoning~\citep{db-hallucination}. However, they do
not provide a general construction of decision boundaries for
autoregressive sequence generation, nor a finite-sampling error theory
for such construction. Our work addresses this gap by defining DPS for
LLM sequence decisions and deriving a practical $K$-sample
approximation with explicit error guarantees. A more detailed related
work discussion is provided in Appendix~\ref{sec:related-appendix}.


%% file: method.tex
\section{Decision Boundary of Language Models}

\subsection{Decision Boundary on Classification Models}
We begin our theoretical analysis with traditional classification
models and aim to extend the insights to
generative language models.

Consider a neural network $f: \mathbb{R}^d \to \mathbb{R}^M$ that
maps an input sample $\mathbf{x} \in \mathbb{R}^d$ to a predicted
probability distribution over $M$ classes, where the set of classes
can be denoted as $\mathcal{M} = \{1, 2,
\dots, M\}$, with $M>2$. Our goal is to characterize the decision boundary of $f$
under a specific input data distribution $\mathcal{D} \subseteq
\mathbb{R}^d$. Without loss of generality, we decompose $f$ into
three components: \emph{(i)} A representation module $h = f_r(\mathbf{x}):
\mathbb{R}^d \to \mathbb{R}^{d'}$ that maps the input
$\mathbf{x}$ to a latent representation $h$.
\emph{(ii)} A linear classification head $z = f_{\text{cls}}(h) =
W_{\text{cls}} h + b_{\text{cls}}: \mathbb{R}^{d'} \to
\mathbb{R}^M$, where $W_{\text{cls}} \in \mathbb{R}^{M \times
d'}$ and $b_{\text{cls}} \in \mathbb{R}^M$ are learnable
parameters, projecting the representation $h$ into
classification logits $z$.
\emph{(iii)} A nonlinear normalization function $P = \sigma(z):
\mathbb{R}^M \to \mathbb{R}^M$, which transforms the logits
into a probability distribution $P = [p_1, p_2, \dots, p_M]$,
where $0 \leq p_i \leq 1$ for $i = 1, \dots, M$ and
$\sum_{i=1}^M p_i = 1$. The final predicted class for
$\mathbf{x}$ can be determined by $\arg\max_i p_i$.
Then, the decision boundary of the neural network $f$ is defined as follows.

\begin{definition}[Decision Boundary of $f$]\label{def:d-b}
The decision boundary of a neural network $f$ under an input
distribution $\mathbf{x}\in\mathcal{D}$ is the set of inputs for which at least two classes in $\mathcal{M} = \{1,
2, \dots, M\}$ have equal and maximal prediction probabilities. Formally, we denote this
set as $\mathcal{B}_M^{(f, \mathcal{D})}$, defined by:
\begin{equation}
\label{eq:d-b}
\begin{aligned}
&\mathcal{B}_M^{(f, \mathcal{D})} = \{ \mathbf{x} \in \mathcal{D}
  \mid \exists~m, n \in \mathcal{M}, m \neq n,\\&\text{ such that } p_m
  = p_n \text{ and } p_m \geq \max_{o \in
    \mathcal{M}\setminus\{m,n\}} p_o \},
\end{aligned}
\end{equation}
where $p_i = P[i] = \sigma(f_{\text{cls}}(f_r(\mathbf{x})))[i]$ is
the predicted probability for class $i$.
\end{definition}

Based on Definition \ref{def:d-b}, we characterize the decision
boundary for multi-class classification scenarios as
follows.

\begin{theorem}[Properties of Multi-Class Classification Boundary]\label{th:mcls}
For multi-class classification ($M > 2$), the decision boundary of
$f$ can be expressed as:
\begin{equation}\small
\label{eq:mcls-b}
\begin{aligned}
&\mathcal{B}_M^{(f,\mathcal{D})} = \bigcup_{1 \leq m < n \leq M} \mathcal{B}_{mn}, \\
&\mathcal{B}_{mn} = \{ \mathbf{x} \mid (w_m - w_n) h +
  (b_m - b_n) = 0, \\&\quad\quad z_m = z_n \geq z_o \, \forall o \neq m, n, \, h
  = f_r(\mathbf{x}), \, \mathbf{x} \in \mathcal{D} \}.
\end{aligned}
\end{equation}
where $z=W_{cls}h+b_{cls}$ is the logits, $w_m$ and $w_n$ are the $m$-th and $n$-th rows of
$W_{\text{cls}}$, and $b_m, b_n$ are the corresponding entries of
$b_{\text{cls}}$.

Geometrically, $\mathcal{B}_M$ induces a Voronoi
partition of the representation space, where each class (i.e.,
decision regions) corresponds to
a Voronoi cell.
\end{theorem}

The proof of Theorem \ref{th:mcls} is
provided in Appendix \ref{sec:proof-mcls}.

\subsection{Decision Boundary for Language Models}

An LLM $f: \mathcal{V}^{N_q} \to \mathcal{V}^{N_r}$ generates a
sequence of tokens $\mathbf{y} = [y_1, \dots, y_{N_r}]$, where each
token $y_t \in \mathcal{V} = \{1, 2, \dots, V\}$ is drawn from a
vocabulary of size $V$, conditioned on an input prompt $\mathbf{x}
= [x_1, \dots, x_{N_q}] \in \mathcal{V}^{N_q}$. $N_{q}$ and $N_{r}$
are the sequence lengths of the input and generated texts. At each generation
step $t$, the LLM predicts the next token $y_t$ based on the
prompt and previously generated tokens, i.e., $y_t \sim P_f(y_t |
\mathbf{x}, y_1, \dots, y_{t-1})$. This single-step generation can be viewed as a
multi-class classification over $\mathcal{V}$, and thus, the
single-token decision boundary follows Theorem \ref{th:mcls}.
When defining the decision boundary for the entire sequence $\mathbf{y}
\in \mathcal{V}^{N_r}$, we first model the joint probability of the
sequence under the autoregressive process. We derive the decision
boundary of LLMs from that of multi-classification, as shown below.

\begin{theorem}[Decision Boundary of Language Models]\label{th:llm-d-b}
The decision boundary of an LLM $f$ under an input text distribution
$\mathcal{D}' \subseteq \bigcup_{n_q=1}^{N_q} \mathcal{V}^{n_q}$ is
the set of prompts $\mathbf{x} \in \mathcal{D}'$ that lead to \textbf{equal
generation probabilities} for at least two distinct sequences
$\mathbf{y}_v, \mathbf{y}_w \in \mathcal{V}^{N_r}$, with their
probabilities being maximal. Formally, the
decision boundary $\mathcal{B}_{llm}^{(f, \mathcal{D}')}$ is:
\begin{equation}\small
\label{eq:d-b-llm}
\begin{aligned}
&\mathcal{B}_{\text{llm}}^{(f, \mathcal{D}')} = \bigcup_{\mathbf{y}_v \neq
                                        \mathbf{y}_w \in
                                        \mathcal{V}^{N_r}}
                                        \mathcal{B}_{\text{llm}, vw},~~ \mathcal{B}_{\text{llm}, vw}= \{ \mathbf{x} \in \mathcal{D}' \mid\\&
                        P_f(\mathbf{y}_v | \mathbf{x}) =
                        P_f(\mathbf{y}_w | \mathbf{x}) \geq
                        \max_{\mathbf{y}_u \in
\mathcal{V}^{N_r}\setminus \{\mathbf{y}_v, \mathbf{y}_w\}}
                        P_f(\mathbf{y}_u | \mathbf{x}) \},
\end{aligned}
\end{equation}
where $P_f(\mathbf{y} | \mathbf{x}) = \prod_{t=1}^{N_r} P_f(y_t |
\mathbf{x}, y_1, \dots, y_{t-1})$ is the joint probability of
generating sequence $\mathbf{y}$ given prompt $\mathbf{x}$.
\end{theorem}

The proof of Theorem \ref{th:llm-d-b} is provided in Appendix
\ref{sec:appendix-llm-d-b}.

While Theorem \ref{th:llm-d-b} provides a concise and intuitive
definition of decision boundary for LLMs,
analyzing or computing this boundary could be computationally impossible in
practice. As analyzed in Section \ref{sec:intro}, the primary
challenge stems from the large vocabulary size and the autoregressive
nature of sequence generation, i.e., for a generation of length
$N_r$, the total number of possible sequences is $V^{N_r}$,
leading to an \emph{exponential growth} of decision regions. Specifically, the decision boundary defined in Equation
\eqref{eq:d-b-llm} involves comparing
$\mathbf{y}_v, \mathbf{y}_w \in \mathcal{V}^{N_r}$, resulting in up
to $\binom{V^{N_r}}{2} \approx \frac{(V^{N_r})^2}{2}$ pairwise
comparisons. This is neither computationally feasible nor
interpretable in subsequent visualizations.

Given the intractability of directly analyzing the decision boundary
defined in Theorem \ref{th:llm-d-b}, a new strategy for constructing
the decision boundary of large language models is
essential. Specifically, this new construction should satisfy
the following criteria:
First, it must be \emph{\textbf{theoretically rigorous}}, meaning the
construction should be equivalent to or provide a bounded
approximation of the decision boundary defined in Theorem
\ref{th:llm-d-b}, ensuring consistency with the formal definition of
the boundary separating prompts that yield different output
sequences. Second, the method should be
\emph{\textbf{practical}}, meaning it must be computationally efficient and
feasible for implementation, enabling the modeling of decision
boundaries for industrial-scale LLMs with large vocabularies and long
generation lengths. Third, the method should be
\emph{\textbf{interpretable}}, meaning the constructed decision boundary
should explicitly capture key properties of LLMs (e.g., curvature),
and provide interpretable insights
into phenomena observed in LLM behavior, such as output variability or
robustness.

In the next section, we will introduce an approximation procedure for
the decision boundary defined in Theorem \ref{th:llm-d-b}, addressing
these criteria to enable practical and meaningful analysis of LLMs.


%% file: method2.tex
\section{K-Grained Decision Potential Surface}

In this section, we introduce the \emph{Decision Potential Surface (DPS)}, a
novel concept for analyzing the decision boundaries of LLMs by
representing the \emph{decision potential} of generated
sequences as a surface over the input manifold. In
Section~\ref{sec:dps}, we formally define DPS and establish its
relationship with the standard decision boundary formulation in
LLMs. In Section~\ref{sec:k-dps}, we propose $K$-grained DPS ($K$-DPS), a
practical approximation of DPS, and theoretically derive its error
bounds with respect to the ideal DPS.

\subsection{Decision Potential Surface of LLMs}\label{sec:dps}
\begin{definition}[Decision Potential Surface of Language Models]\label{def:ds}
Given an input text distribution $\mathbf{x} \in \mathcal{D}'$ with
$\mathcal{D}' \subseteq \bigcup_{n_q=1}^{N_q} \mathcal{V}^{n_q}$ and
a language model $f: \mathcal{V}^{N_q} \to \mathcal{V}^{N_r}$ that
generates an output sequence $\mathbf{y} = f(\mathbf{x}) =
\arg\max_{\mathbf{y}_s \in \mathcal{V}^{N_r}} P_f(\mathbf{y}_s |
\mathbf{x})$, we define the \emph{decision
  potential function (DPF)} $\Phi_f^{\infty}(\mathbf{x}):
\mathcal{D}' \to \mathbb{R}_+$ as the squared difference in
log-likelihoods between the top two generated sequences under the
input prompt $\mathbf{x}$, i.e.,
\begin{equation}\small
\label{eq:phi}
\begin{aligned}
&\Phi_f^{\infty}(\mathbf{x})\\&= \left( \min_{\mathbf{y}_w \in
                              \mathcal{V}^{N_r}, \mathbf{y}_w \neq
                              \mathbf{y}_v} \left[ \max_{\mathbf{y}_v
                              \in \mathcal{V}^{N_r}} \log
                              P_f(\mathbf{y}_v | \mathbf{x}) - \log
                              P_f(\mathbf{y}_w | \mathbf{x}) \right]
                              \right)^2 \\
&= \big( \log P_f(\mathbf{y}_{1*} | \mathbf{x}) - \log P_f(\mathbf{y}_{2*} |
  \mathbf{x}) \big)^2,
\end{aligned}
\end{equation}
where $\mathbf{y}_{1*}, \mathbf{y}_{2*} \in \mathcal{V}^{N_r}$ denote the
sequences with the highest and second-highest log-likelihoods,
respectively. The \emph{decision potential surface (DPS)} is then
defined as $\mathcal{S}^{(f, \mathcal{D}')} := \{
\Phi_f^{\infty}(\mathbf{x}) \mid \mathbf{x} \in \mathcal{D}' \}$.
\end{definition}

Intuitively, $\mathcal{S}^{(f, \mathcal{D}')}$ can be viewed as a
surface representing the competitive likelihoods across all
inputs, where each decision potential value
$\Phi_f^{\infty}(\mathbf{x})$ quantifies the \emph{confidence} in
distinguishing the most likely sequence.

\begin{remark}[The Top-1/Top-2 Construction Is Naturally Multiclass, Not Binary]\label{rem:multiclass}
The use of only the top two sequences in Definition \ref{def:ds} follows the
classical definition of multiclass decision boundaries. A multiclass
boundary is not obtained by comparing all classes simultaneously at
every input point; rather, as established in
Theorem~\ref{th:mcls}, it is the union of local pairwise boundaries
where the \emph{two maximal} classes tie. In the LLM setting, the
``classes'' are possible generated sequences or token continuations.
Therefore, \textbf{restricting attention to the top-1 and top-2 sequences is
not a binary approximation but the standard local
characterization} of a multiclass decision boundary.
\end{remark}

Following Definition \ref{def:ds}, we define \emph{isohypses}
(i.e., contour lines) on the surface $\mathcal{S}^{(f,\mathcal{D}')}$ as follows:
\begin{definition}[$\varepsilon$-Isohypse]\label{def:epsilon-isohypse}
The $\varepsilon$-isohypse on the decision potential surface
$\mathcal{S}^{(f, \mathcal{D}')}$ is the set of inputs with the
same decision potential value $\varepsilon$, i.e.,
\begin{equation}
\label{eq:3}
\mathcal{D}'_{(\varepsilon, f)} = \{ \mathbf{x} \mid \mathbf{x} \in \mathcal{D}';
\Phi_f^{\infty}(\mathbf{x}) = \varepsilon \}.
\end{equation}
\end{definition}

As a degenerate case, the zero level set of $\mathcal{S}^{(f,
  \mathcal{D}')}$ exhibits the following
property:
\begin{theorem}[0-Isohypse as the Decision Boundary]\label{th:0-isohypse}
The decision boundary of a language model $f(\mathbf{x})$ under $\mathcal{D}'$, as defined in Theorem \ref{th:llm-d-b}, is equivalent to the 0-isohypse, i.e.,
\begin{equation}
\label{eq:isohypse}
\mathcal{B}_{llm}^{(f, \mathcal{D}')} = \mathcal{D}'_{(0,f)} = \{ \mathbf{x} \in \mathcal{D}' \mid \Phi_f^{\infty}(\mathbf{x}) = 0 \},
\end{equation}
where regions separated by the 0-isohypse correspond exactly to
the Voronoi cells.
\end{theorem}

We also provide the following corollary to characterize the surface structure:
\begin{corollary}[$\varepsilon$-Isohypse Gives 
  $\sqrt{\varepsilon}$-nat Confidence Hierarchy]\label{cor:epsilon-confidence}
For any $\varepsilon > 0$, the input space $\mathcal{D}'$ is partitioned into three disjoint strata:
\begin{itemize}
    \item $\varepsilon$-barrier:
      $\mathcal{D}'_{(>\varepsilon,f)}=\{\mathbf{x}|\Phi_{f}^{\infty}(\mathbf{x})>\varepsilon;\mathbf{x}\in
    \mathcal{D}'\}$, where $f(\mathbf{x})$ predicts the sequence of
    its region with at least $\sqrt{\varepsilon}$ nats (natural units
    of information) of confidence over the next most likely sequence.
    \item $\varepsilon$-well:
      $\mathcal{D}'_{(<\varepsilon,f)}=\{\mathbf{x}|\Phi_{f}^{\infty}(\mathbf{x})<\varepsilon;\mathbf{x}\in
    \mathcal{D}'\}$, where $f(\mathbf{x})$ has low confidence, with
    a margin less than $\sqrt{\varepsilon}$ nats. As $\varepsilon \to
    0$, this stratum converges to the 0-isohypse.
    \item $\varepsilon$-isohypse: $\mathcal{D}'_{(\varepsilon,f)} = \{
      \mathbf{x} \in \mathcal{D}' \mid \Phi_f^{\infty}(\mathbf{x}) =
      \varepsilon \}$, representing the contour where the confidence 
      margin is exactly $\sqrt{\varepsilon}$ nats.
\end{itemize}
\end{corollary}

Proofs are provided in Appendix \ref{proof-th:0-isohypse} and
\ref{proof-cor:epsilon-confidence}, respectively.

Given Theorem \ref{th:0-isohypse}, we can construct the DPS
defined in Definition \ref{def:ds} to characterize the decision
boundaries of LLMs.
Unfortunately, computing the decision boundary or visualizing the
DPS based on Definition \ref{def:ds} remains computationally infeasible, as evaluating
$\Phi_f^{\infty}(\mathbf{x})$ in Equation \eqref{eq:phi} requires
considering all possible sequences in $\mathcal{V}^{N_r}$, resulting
in a computational complexity the same as before.

Fortunately, as Equation \eqref{eq:phi} depends only on the
log-likelihoods of the top two sequences, we can propose an efficient
approximation with a modest error, detailed in the next subsection.

\subsection{K-Grained Decision Potential Surface}
\label{sec:k-dps}

We introduce $K$-grained decision potential surface for approximating
$\mathcal{S}^{(f,\mathcal{D}')}$:
\begin{definition}[$K$-Grained Decision Potential Surface]
  Given $\mathbf{x}\in\mathcal{D}'$ and a language model
  $f(\mathbf{x})$, we define the \emph{$K$-grained potential function}
  $\Phi_{f}^{K}(\mathbf{x}):\mathcal{D}'\rightarrow \mathbb{R}_{+}$ as
  \begin{equation}\small
  \label{eq:k-dpf}
  \begin{aligned}
 &\Phi_{f}^{K}(\mathbf{x})\\&=\left(\min_{\mathbf{y}_{w}\in\mathcal{Y}_{K},\mathbf{y}_{w}\neq\mathbf{y}_{v}}{\left[\max_{\mathbf{y}_{v}\in\mathcal{Y}_{K}}{\log
       P_{f}(\mathbf{y}_{v}|\mathbf{x})}-\log P_{f}(\mathbf{y}_{w}|\mathbf{x})\right]}\right)^{2}\\
                          &=\big(\log {P}_{f}(\mathbf{y}_{1*}^{K}|\mathbf{x})-\log {P}_{f}(\mathbf{y}_{2*}^{K}|\mathbf{x})\big)^{2},
  \end{aligned}
  \end{equation}
  where $1\ll K\ll V^{N_{r}}$ denotes the size of output space for
  each input, $\mathcal{Y}_{K}=\{\mathbf{y}_{v}\sim
  P_{f}(\cdot|\mathbf{x})|v=1,...,K\}$ denotes $K$ \emph{i.i.d. (independent and identically distributed)} sampled
  texts, and $\mathbf{y}_{1*}^{K}$ and $\mathbf{y}_{2*}^{K}$ denote the
  top two generated texts with the largest generation log-likelihoods
  within $\mathcal{Y}_{K}$.
\end{definition}

In this way, the computational complexity of constructing the decision boundary is reduced from $\mathcal{O}(V^{2N_{r}}\cdot|\mathcal{D}'|)$ to $\mathcal{O}(K^{2}\cdot |\mathcal{D}'|
)$, resulting in a substantial reduction. This naturally leads to the
next question: what is the error between $\Phi_{f}^{K}(\mathbf{x})$ and
$\Phi_{f}^{\infty}(\mathbf{x})$? We address this by theoretically
analyzing their relationship in the following theorems.

\begin{theorem}[Error Bound for Estimating $\Phi_f^\infty(\mathbf{x})$ with $\Phi_f^K(\mathbf{x})$]\label{th:error}
For a fixed input $\mathbf{x} \in \mathcal{D}'$ and a set
$\mathcal{Y}_K$ of $K$ i.i.d. samples drawn from the language model's
output distribution $P_f(\cdot | \mathbf{x})$, suppose the population
top-2 gap satisfies $\Delta_\infty(\mathbf{x}) = \log
P_f(\mathbf{y}_{1*} | \mathbf{x}) - \log P_f(\mathbf{y}_{2*} |
\mathbf{x}) \leq R_K(\mathbf{x})$, where
$R_K(\mathbf{x}) = \log P_f(\mathbf{y}_{1*}^K | \mathbf{x}) -
\min_{\mathbf{y} \in \mathcal{Y}_K} \log P_f(\mathbf{y} | \mathbf{x})$
represents the log-likelihood diameter of $\mathcal{Y}_K$. Then, for
any $\delta \in (0,1)$, the error between the sample-based decision
potential $\Phi_f^K(\mathbf{x})$ and the true decision potential
$\Phi_f^\infty(\mathbf{x})$ satisfies:
\begin{equation}
\label{eq:absolute-error}
|\Phi_f^K(\mathbf{x}) - \Phi_f^\infty(\mathbf{x})| \leq 2 R_K^2(\mathbf{x}) \sqrt{\frac{\log(4/\delta)}{2K}},
\end{equation}
with probability at least $1 - \delta - 2\varepsilon_{\text{tail}}$,
where $\varepsilon_{\text{tail}} = \left(1 - P_f(\mathbf{y}_{1*}^K |
  \mathbf{x})\right)^K$.
\end{theorem}

\begin{theorem}[Local Candidate-Set Bound]\label{th:local-candidate-bound}
For a fixed input $\mathbf{x}\in\mathcal{D}'$, let
$\Delta_\infty(\mathbf{x}) = \log P_f(\mathbf{y}_{1*}|\mathbf{x}) -
\log P_f(\mathbf{y}_{2*}|\mathbf{x})$. For any $\eta\geq 0$, define
the local near-second candidate set
\begin{equation}
\label{eq:local-second-set}
\begin{aligned}
\mathcal{Y}_{2,\eta}(\mathbf{x})=\{&
\mathbf{y}\in\mathcal{V}^{N_r}: \mathbf{y}\neq\mathbf{y}_{1*},\\
&\log P_f(\mathbf{y}|\mathbf{x})
\geq \log P_f(\mathbf{y}_{2*}|\mathbf{x})-\eta\}.
\end{aligned}
\end{equation}
Let
\begin{equation}
\label{eq:local-second-mass}
p_{2,\eta}(\mathbf{x})=
\sum_{\mathbf{y}\in\mathcal{Y}_{2,\eta}(\mathbf{x})}
P_f(\mathbf{y}|\mathbf{x})
\end{equation}
be the probability mass of this local candidate set. Then
\begin{equation}
\label{eq:local-candidate-bound}
|\Phi_f^K(\mathbf{x})-\Phi_f^\infty(\mathbf{x})|
\leq \eta\left(2\Delta_\infty(\mathbf{x})+\eta\right)
\end{equation}
with probability at least
\begin{equation}
\label{eq:local-candidate-prob}
1-\left(1-P_f(\mathbf{y}_{1*}|\mathbf{x})\right)^K
-\left(1-p_{2,\eta}(\mathbf{x})\right)^K .
\end{equation}
Equivalently, when $p_{2,\eta}(\mathbf{x})>0$, if
\begin{equation}
K\geq \max\left\{
\frac{\log(2/\delta)}{P_f(\mathbf{y}_{1*}|\mathbf{x})},
\frac{\log(2/\delta)}{p_{2,\eta}(\mathbf{x})}
\right\},
\end{equation}
then
Equation~\eqref{eq:local-candidate-bound} holds with probability at
least $1-\delta$.
\end{theorem}

\begin{theorem}[Expected Error Bound]\label{th:e-error}
Under the same conditions as Theorem \ref{th:error}, the expected error
between the sample-based decision potential $\Phi_f^K(\mathbf{x})$ and
the true decision potential $\Phi_f^\infty(\mathbf{x})$ is bounded as:
\begin{equation}
\label{eq:33}
\mathbb{E}\left[|\Phi_f^K(\mathbf{x}) - \Phi_f^\infty(\mathbf{x})|\right] \leq 2 R_K^2(\mathbf{x}) \sqrt{\frac{2\pi}{K}} + 4 R_K^2(\mathbf{x}) \varepsilon_{\text{tail}},
\end{equation}
where $\varepsilon_{\text{tail}} = \left(1 - P_f(\mathbf{y}_{1*}^K | \mathbf{x})\right)^K$.
\end{theorem}

\begin{figure*}[t]
\centering
\begin{minipage}{0.49\linewidth}
\centering
\includegraphics[width=\linewidth,height=0.145\textheight,keepaspectratio]{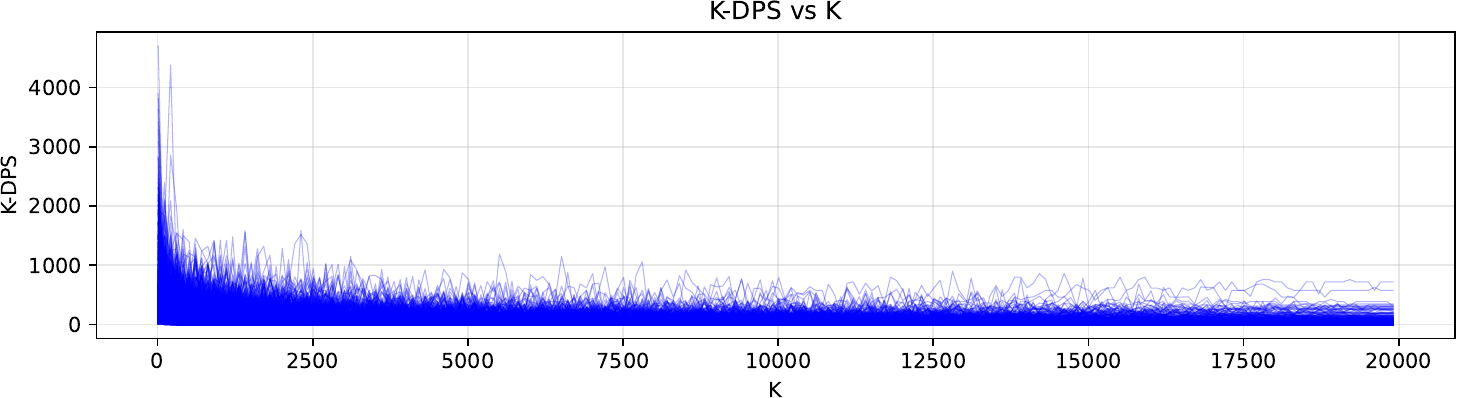}
\caption{Effect of sampling size $K$ on the $K$-DPS value.}
\label{fig:varyk-kdps}
\end{minipage}
\hfill
\begin{minipage}{0.49\linewidth}
\centering
\includegraphics[width=\linewidth,height=0.145\textheight,keepaspectratio]{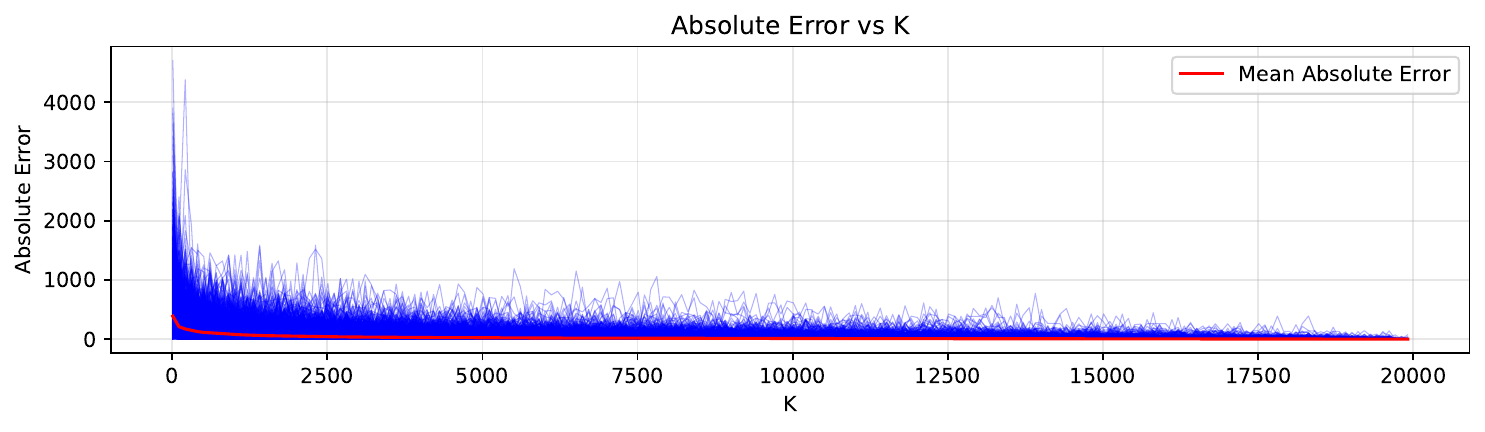}
\caption{Absolute error between the high-budget reference
$K$-DPS ($K=20{,}000$) and estimates with varying $K$.}
\label{fig:varyk-error}
\end{minipage}
\end{figure*}

\begin{figure*}[t]
\centering
\includegraphics[width=0.999\linewidth,height=0.16\textheight,keepaspectratio]{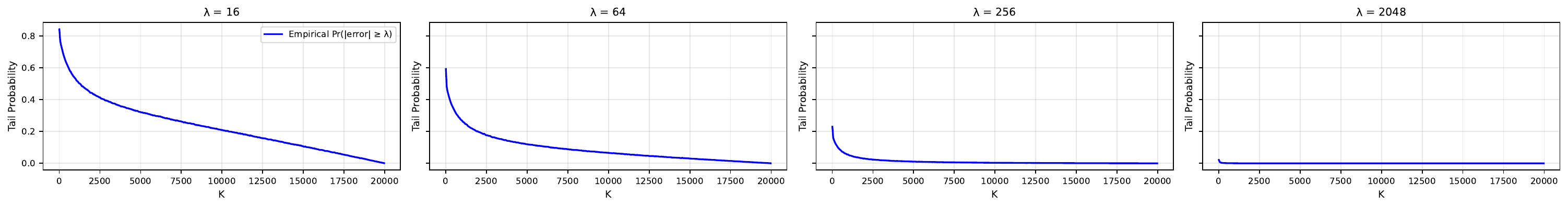}
\caption{Empirical concentration experiments with different $\lambda$ values.}
\label{fig:concentration}
\end{figure*}

\begin{corollary}[Concentration Bound]\label{th:tail-error}
Under the same conditions as Theorem \ref{th:error}, for any $\lambda
> 0$, the tail probability of the error satisfies:
\begin{equation}\small
\label{eq:34}
\Pr\left(|\Phi_f^K(\mathbf{x}) - \Phi_f^\infty(\mathbf{x})| \geq \lambda\right) \leq 4 \exp\left(-\frac{K \lambda^2}{2 R_K^4(\mathbf{x})}\right) + 2 \varepsilon_{\text{tail}},
\end{equation}
where $\varepsilon_{\text{tail}} = \left(1 - P_f(\mathbf{y}_{1*}^K | \mathbf{x})\right)^K$.
\end{corollary}

Proofs are provided in Appendix \ref{proof-th:error}, Appendix
\ref{proof-th:local-candidate-bound}, Appendix \ref{proof-th:e-error},
and Appendix \ref{proof-th:tail-error}, respectively.

The estimation error contracts with $K$ at the familiar $1/\sqrt{K}$
rate, mirroring the decay of an empirical mean. The tail probability
$\varepsilon_{\text{tail}}$ bounds the chance that the true top-1
candidate is absent from $\mathcal{Y}_K$. This term depends on the
concentration of the output distribution and is exponentially
suppressed as $K$ grows. In practice, its magnitude differs across
generation regimes:
\emph{i) Ordinary generation.} When $N_r$ is small, the
top sequences carry substantial joint probability mass, and
$\varepsilon_{\text{tail}}$ becomes negligible at modest $K$ (e.g.,
$K \approx 10^3$). Token-level DPS ($N_r=1$, Section~\ref{sec:token-dps})
is even more favorable, as the output space is only the vocabulary.
\emph{ii) Long or high-entropy generation.} Very long, high-entropy generation
can produce tiny joint probabilities for any individual sequence.
This is an inherent difficulty of sequence-level decision-boundary
construction for LLMs, not a defect specific to DPS. In such regimes,
one may instead rely on prefix-conditioned token-level DPS or report
empirical stability curves (Section~\ref{sec:impli}) rather than
relying merely on the joint-probability tail term.

The common factor $R_K(\mathbf{x})$ is a worst-case log-likelihood
diameter dictated by the least likely sentence that happens to be
sampled. Theorem~\ref{th:local-candidate-bound} gives a sharper
local alternative: the error depends on the tolerance $\eta$ around
the true second-best candidate and on the probability mass
$p_{2,\eta}(\mathbf{x})$ of candidates in that local band. When $\eta=0$,
the bound becomes exact: if the sample contains the true top-1
sequence and any true second-best sequence, then
$\Phi_f^K(\mathbf{x})=\Phi_f^\infty(\mathbf{x})$. For larger
$\eta$, the theorem separates candidate discovery
($P_f(\mathbf{y}_{1*}|\mathbf{x})$ and $p_{2,\eta}(\mathbf{x})$) from
gap distortion ($\eta(2\Delta_\infty+\eta)$), avoiding dependence on
the least likely sampled sentence.

\subsection{Variants of $K$-DPS}

\noindent
\textbf{Semantically Similar Text Completions.}\label{sec:deg}
One may ask whether the top-1 and top-2 completions are often nearly
identical, which could inflate DPS values artificially and obscure
genuine decision boundaries. DPS naturally accommodates this scenario.
On the one hand, DPS is defined on the raw
sequence/token probability space of the language model, where semantic
similarity between completions is not ignored but naturally reflected
in the geometry. On the other hand, semantically equivalent outputs
occupy adjacent or overlapping decision regions, producing narrow
basins that match the intuition of a smooth semantic equivalence
class, and this clustering is an emergent feature of the multiclass
decision boundary.
An empirical validation of this point is provided in
Appendix (Table \ref{tab:edit-dis}). We further discuss how to filter
similar texts when desired and why such filtering
does not make the theoretical guarantees depend on the worst-case
diameter $R_K(\mathbf{x})$ in Appendix~\ref{sec:filter-bound}.

\noindent
\textbf{Influence of Sampling Strategies.}\label{sec:temperature}
Sampling temperature and other decoding strategies change the output
distribution and thus the numerical values of the DPS, without
invalidating the framework. The DPS is defined on the model's raw
output distribution (Definition~\ref{def:ds}), raising a natural
question: is it compatible with different decoding strategies?

The answer is affirmative, though the nature of compatibility varies
by strategy. As proved in Appendix~\ref{sec:T}, temperature scaling
constitutes a monotone transformation of the DPS: for any temperatures
$T_1<T_2$, the ordering of output sequences by log-probability is
preserved, so the zero-height isohypse $\mathcal{D}'_{(0,f)}$ and the
decision boundary structure remain identical. Higher temperature
compresses the surface vertically by a factor of $1/T^2$, reducing
the dynamic range of DPS values, while lower temperature amplifies
local contrast. All topological features, including the arrangement of
isohypses, are strictly preserved.

Other decoding strategies require separate treatment, as discussed in
Appendix~\ref{sec:sampling}. Nucleus (top-$p$) sampling restricts
candidates to a high-probability subset of the model's support,
leaving the error bounds of Theorems~\ref{th:error}--\ref{th:tail-error}
valid with slightly adjusted tail constants for typical
$p\in[0.9,1.0)$. Top-$k$ sampling, when $k$ is small, alters the
sampling distribution more substantially and requires modified
theoretical bounds; we recommend $k\geq 50$ so that the top tokens at
each step are reliably included, which suffices for accurate DPS
estimation at modest $K$.

\noindent
\textbf{Token-Level DPS.}\label{sec:token-dps}
While the preceding sections focus on sequence-level decision boundary
construction, a degenerate case, namely token-level DPS, merits
separate discussion.
The sequence-level definition in Equation~\eqref{eq:phi} naturally supports
token-level analysis as the special case $N_r = 1$.
For a generation step $t$, the input consists of the prompt
concatenated with the previously generated prefix
$\mathbf{x}' = [\mathbf{x}, y_1, \dots, y_{t-1}]$, and the
output classes are the vocabulary tokens $\mathcal{V}$. The decision
potential function reduces to the squared log-likelihood
difference between the top two tokens, while all theoretical guarantees
(Theorems~\ref{th:error}--\ref{th:tail-error}) carry over with the
output space being only $V$ categories.


%% file: exper.tex
\section{Empirical Analysis}


\subsection{Settings}
\noindent
\textbf{Datasets and Models.} 
We utilize both pre-training corpora and supervised fine-tuning (SFT)
datasets to simulate the input data distribution for constructing
decision boundaries and the decision potential surface.
For the
pre-training corpus, we select Wikipedia Mini~\citep{wiki-mini}, an
unsupervised text corpus containing a condensed version of Wikipedia
articles. For supervised fine-tuning, we employ
Tulu-3-SFT-MIX~\citep{tulu3-sft}, OpenO1-SFT~\citep{openo1},
HH-RLHF~\citep{hh-rlhf}, and Alpaca~\citep{alpaca}, all of which are
widely used in academic and industrial settings.
We use Llama3.2-1B~\citep{llama3} as the basic backbone, and employ 
Llama-3.1 (8B)~\citep{llama3}, Llama-Guard-3 (8B)~\citep{llama3}, Mistral
(7B)~\citep{DBLP:journals/corr/abs-2310-06825}, Zephyr (7B)~\cite{DBLP:journals/corr/abs-2310-16944}, and Tulu-2 (7B)~\cite{DBLP:journals/corr/abs-2311-10702} for the alignment decision boundary analysis, and utilize
the Llama-2 (7B)~\cite{DBLP:journals/corr/abs-2307-09288} for the machine unlearning experiments.

\noindent
\textbf{Implementation Details.} 
For sampling, we
use \emph{nucleus sampling} in our $K$-DPS implementation, with the
clipping probability $p$ set to 0.9. In subsequent experiments, each
data point is repeated five times.
The experiments are conducted on
$4~\times$ 94GB Nvidia Tesla H100 NVL GPUs.

\begin{figure*}[h]
\centering
\includegraphics[width=0.99\linewidth,height=0.18\textheight,keepaspectratio]{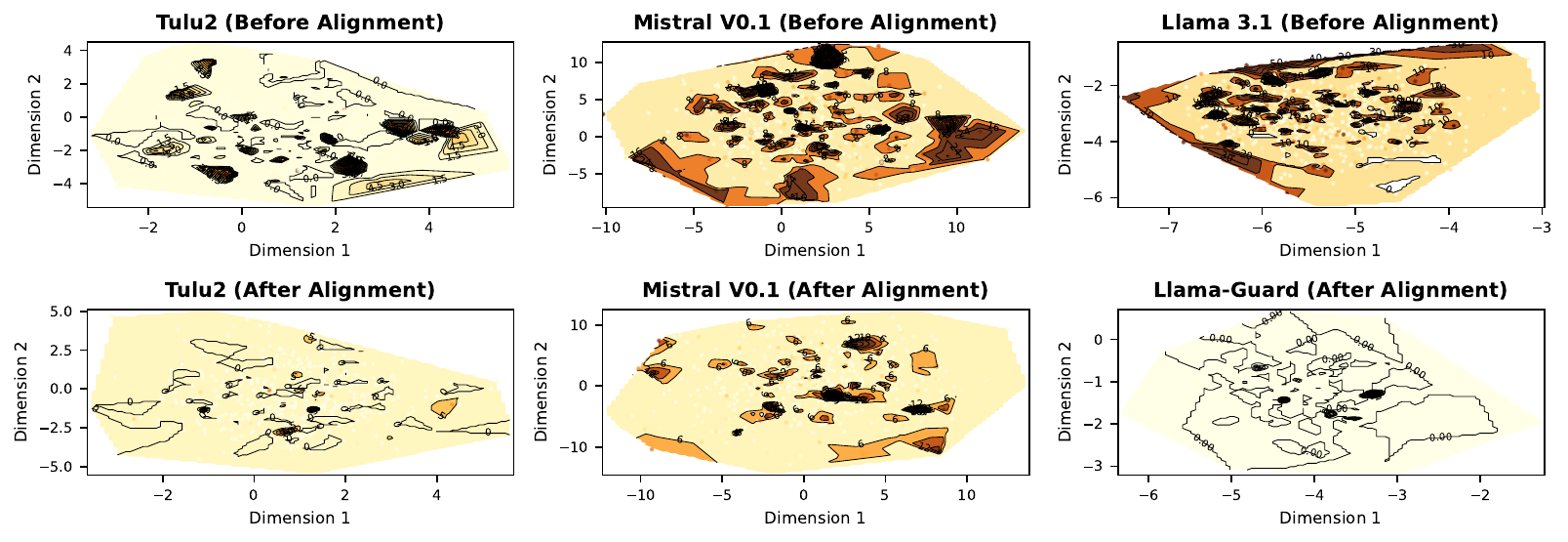}
\caption{Comparison of $K$-DPS for models before and after
  alignment. A deeper color indicates a higher $K$-DPS score.}
\label{fig:vis-alignment}
\end{figure*}

\begin{figure*}[h]
\centering
\includegraphics[width=0.99\linewidth,height=0.19\textheight,keepaspectratio]{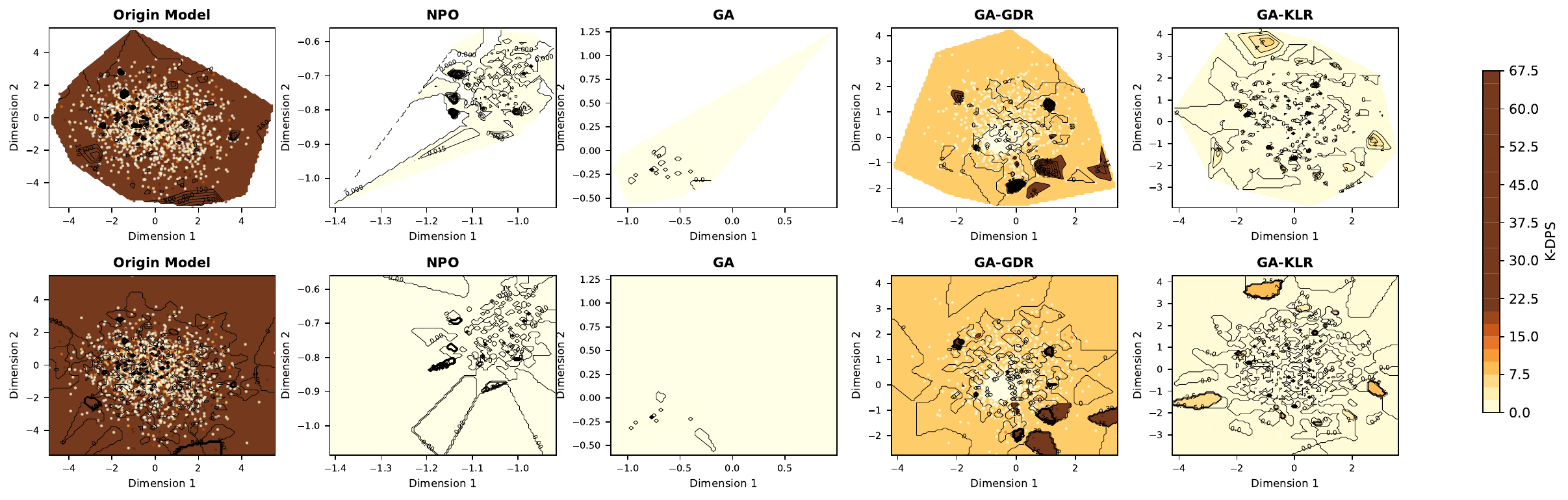}
\caption{Comparison of $K$-DPS among different machine unlearning
  algorithms. The first and second rows indicate the visualization
  under linear and nearest interpolation, respectively.}
\label{fig:vis-unlearn}
\end{figure*}

\subsection{Influence of Sampling Grain $K$}

We first evaluate the impact of the key hyperparameter, the sampling
grain $K$, on the $K$-DPS value and the absolute errors between
$K$-DPS and the ideal DPS. Specifically, we set $K=20,000$ as a
high-budget reference estimator to
approximate the ideal DPS (exact exhaustive enumeration of
$\mathcal{V}^{N_r}$ being computationally infeasible for modern
LLMs). We then compute the $K$-DPS
values by varying $K$ from 10 to 20,000 to illustrate how the decision
potential value $\Phi_{f}^{K}(\cdot)$ converges to the reference
$\Phi_{f}^{20,000}(\cdot)$. Similarly, we calculate the absolute errors of
$\Phi_{f}^{K}(\cdot)$ across different settings of $K$.
As shown in Figure~\ref{fig:varyk-kdps}, the potential values rapidly
converge to the reference values (represented by horizontal lines at the
tails), indicating that a relatively small $K$ can yield a highly
accurate decision potential surface. Moreover, by examining the
errors defined in Equation~\eqref{eq:absolute-error}, as depicted in
Figure~\ref{fig:varyk-error}, we observe that both the absolute error
for individual samples and the empirical average error decrease to
zero, confirming the effectiveness of
$K$-DPS. Figures~\ref{fig:varyk-kdps} and~\ref{fig:varyk-error} also
serve as valuable references for selecting appropriate $K$ values.
More error analysis can be found in the Appendix.

\subsection{Empirical Concentration Bias}

We also present an empirical study of concentration
experiments, focusing on the trend of sample probabilities for inputs
with a decision potential error exceeding a given fixed value
$\lambda$ across various sampling sizes $K$. As shown in
Figure~\ref{fig:concentration}, we evaluate the tail probability for
$K$ values ranging from 10 to 20,000, with $\lambda$ set to 16, 64,
256, and 2048. These $\lambda$ values represent the geometric errors
between the approximate and ideal DPS values. It is noteworthy to
emphasize that even a $\lambda$ value of
256 is not excessively large or insignificant, as our decision
potential function $\Phi_{f}^{K}(\cdot)$ is defined as the
\textbf{square} of logarithmic errors, as specified in
Equation~\ref{eq:k-dpf}.

From Figure~\ref{fig:concentration}, we observe that the tail
probabilities exhibit an exponential decrease, indicating that the
likelihood of exceeding a given error bound diminishes significantly
with a linear increase in the sampling size $K$. Specifically,
Figure~\ref{fig:concentration} demonstrates that a sampling size of
10,000 ensures an absolute error below 64 with 90\% confidence and an
error below 256 with 99\% probability. These results align
closely with our absolute error analysis presented in
Figure~\ref{fig:varyk-error}.

\subsection{Implications}\label{sec:impli}

In this section, we choose two critical topics on LLMs, alignment and
machine unlearning, as proof-of-concept examples to demonstrate the
effectiveness of $K$-DPS for intuitively interpreting LLMs.

\noindent
\textbf{Alignment.}
We use $K=2,500$ for the alignment experiments, 
with input queries from AdvBench~\citep{zou2023universal}.
As shown in Figure \ref{fig:vis-alignment}, the decision boundary of
aligned models becomes dramatically smoother and flatter compared to their
pre-alignment counterparts when evaluated on adversarial prompts from
AdvBench. This indicates that alignment
substantially reduces regions of high confidence in harmful outputs
and creates broad, low-$K$-DPS basins that strongly favor
refusal. This geometric transformation directly explains both:
\emph{i) Why jailbreaks succeed on unaligned models}: they target
narrow, high-confidence ``vulnerability spikes'' that remain in the
pre-alignment landscape;
\emph{ii) Why alignment mitigates most jailbreaks}: it eliminates
these spikes entirely, making harmful responses probabilistically
unlikely across vast regions of prompt space.
Note that this phenomenon is observed from the height of the surface,
which is not affected by dimensionality reduction.


\noindent
\textbf{Machine Unlearning.}
We apply our $K$-DPS to two representative unlearning methods: Gradient
Ascent (GA) \citep{ga} and Negative Preference Optimization (NPO) \citep{npo}. We use
the standard Harry Potter book as the forget (unlearning) corpus and a
Wikipedia subset~\cite{wiki-mini} as the retain set. During unlearning, we continue
training on the retain set using standard gradient descent (GDR) or
the KL divergence (KLR) from the original model. We set $K$ = 2,000.
As shown in Figure \ref{fig:vis-unlearn}, our $K$-DPS visualizations
provide a far clearer picture of the side effects of unlearning than
previously possible \citep{sur1-unlearn,sur2-unlearn}.
While current works could merely report that unlearning without proper retention
training degrades overall performance, they are unable to show what
form this degradation takes in the model's internal decision
process. With $K$-DPS, we reveal that na\"ive unlearning methods (e.g.,
GA) can trigger catastrophic collapse of the
entire decision manifold, where large portions of the prompt space that were
previously smooth become extremely jagged and fragmented, with erratic
high- and low-$K$-DPS spikes appearing in regions unrelated to the
forget corpus.
In contrast, when retention training is included, the damage is substantially mitigated.
The above observations demonstrate that $K$-DPS not only confirms known
phenomena at a qualitative level but, for the first time, makes the
geometric nature of ``machine unlearning damage'' directly observable and
comparable across different methods.



%% file: appendix.tex
\section{LLM Usage Statement}

AI tools were used for error checking, proofreading, result visualization, and code optimization.

\section{Additional Related Work}\label{sec:related-appendix}

\noindent
\textbf{Decision Boundary Analysis on ML Models.}
The earliest exploration of decision boundaries in neural networks
dates back to the era of linear classifiers and shallow
architectures. \citet{rosenblatt1958perceptron} introduced the first
linear decision boundary for binary classification, where a hyperplane
separates input samples into two classes. For shallow feedforward
neural networks (FFNNs) with non-linear activations (e.g., sigmoid,
ReLU), several works~\cite{lee1997decision,turner1996analysis} quantified how hidden layers enable non-linear
decision boundaries. For instance, \citet{lee1997decision} proposed a feature
extraction method that maps input data to a space aligned with FFNN
decision boundaries, showing that boundary curvature correlates with
model capacity and classification accuracy.
The connection between the stability of neural network decision
boundaries and overall error performance
has also been revealed~\citep{turner1996analysis} under ensembling.
In recent years, researchers extended decision boundary analysis to
convolutional neural networks (CNNs) and transformers. Specifically, \citet{goodfellow2015explaining} revealed a key
vulnerability of deep CNNs: their decision boundaries are locally
linear in high-dimensional input spaces, making them susceptible to
adversarial examples. Then, \citet{madry2018towards} further formalized this
by proving that robust training (e.g., adversarial training) ``smooths''
decision boundaries, reducing local linearity and adversarial
susceptibility. Similarly, \citet{gu2017badnets} focused on backdoor
attacks in CNNs, linking them to hidden ``trapdoors'' in decision
boundaries. Such attacks involve planting a small, specific pattern
that shifts the boundary and forces misclassification for triggered
inputs.
\citet{lee1997decision} laid the groundwork by introducing decision boundary
feature extraction and highlighting the role of boundaries in
characterizing network behavior before deep learning. Later,
\citet{db6} examined the decision boundaries of trained networks, analyzing
how
architectural elements (e.g., depth and activation functions) and training data
influence boundary shape, complexity, and stability, providing
insights into network task performance. \citet{db5} conducted an
empirical study on deep network boundaries across CV tasks, including image
classification and object detection. Through quantitative and qualitative
analysis, they explored boundary behavior near correct and misclassified
samples and adversarial examples, bridging theory-practice gaps. Similarly,
\citet{db2} reviewed boundary research challenges such as high input
dimensionality, complex architectures, limited visualization tools,
and opportunities, including advanced math, innovative visualization, and robustness
enhancements. \citet{db1} complementarily proposed metrics
like smoothness, curvature, and class separation to quantify boundaries,
enabling cross-model comparisons and standardized analysis for deep
learning interpretability.

\noindent
\textbf{Decision Boundary Analysis on LLMs.}
Research on LLMs mainly focuses on exploring how this concept
illuminates the decision-making mechanisms and inherent limitations of
LLMs.
As an example, \citet{db-icl} probed the decision boundaries of
in-context learning in LLMs, shedding light on how contextual
information shapes boundary formation and decision outputs. Another
work~\cite{db-memorization} revealed that LLMs lack awareness of their
own decision confidences and that self-generated counterfactual
explanations are unreliable. With respect to reasoning ability,
BARREL~\cite{db-hallucination} designs a boundary-aware reasoning
framework to enhance the factual accuracy of LLMs via boundary
awareness.
However, these preliminary explorations fail to address the core
challenges of LLM decision boundary analysis: how to construct
decision boundaries for the generalized LLM token generation task
(which extends beyond specially designed toy classification tasks) in
a computationally feasible manner? how to theoretically analyze the
construction error for the decision boundary?
To fill this gap, we aim to propose a new decision boundary theory to
address the high-dimensional complexity and construction barriers of
LLMs, enabling accurate, efficient, and interpretable boundary
modeling that aligns with the inherent characteristics of LLMs.

\begin{figure*}[t]
\centering
\includegraphics[width=0.94\linewidth]{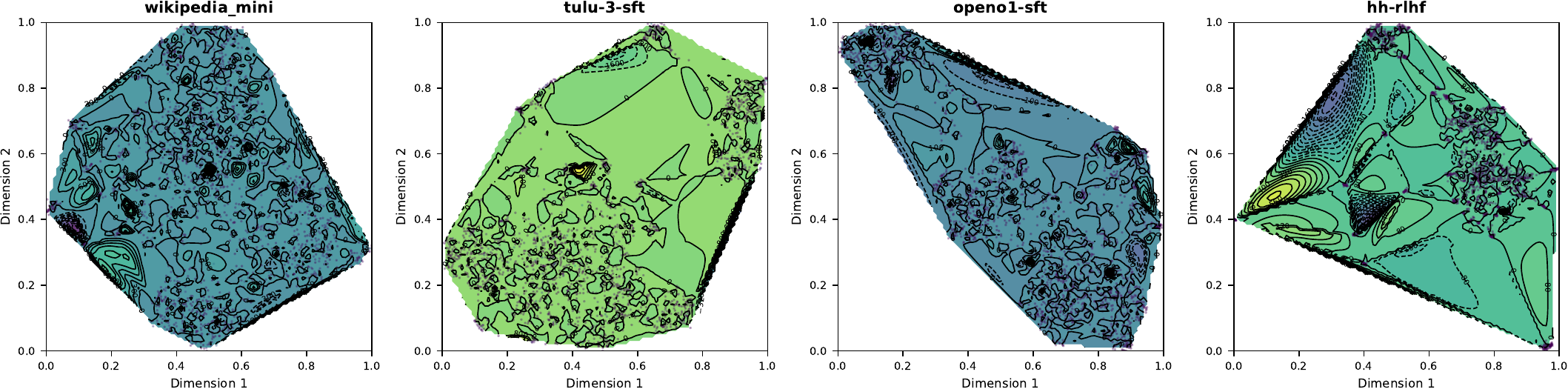}
\caption{Contour visualization of the $K$-DPS ($K=2,500$) for
  Llama-3.2-1B on four datasets. Region colors
  represent the decision potential values. Black lines denote isohypses,
  with the 0-isohypse indicating the decision
  boundary. Cubic interpolation is applied to construct the mesh grid,
  with visualizations using linear and nearest interpolation shown in
  Figures~\ref{fig:vis-contour-linear}
  and~\ref{fig:vis-contour-nearest}.}
\label{fig:vis-contour-cubic}
\end{figure*}

\begin{figure*}[h]
\centering
\includegraphics[width=0.94\linewidth]{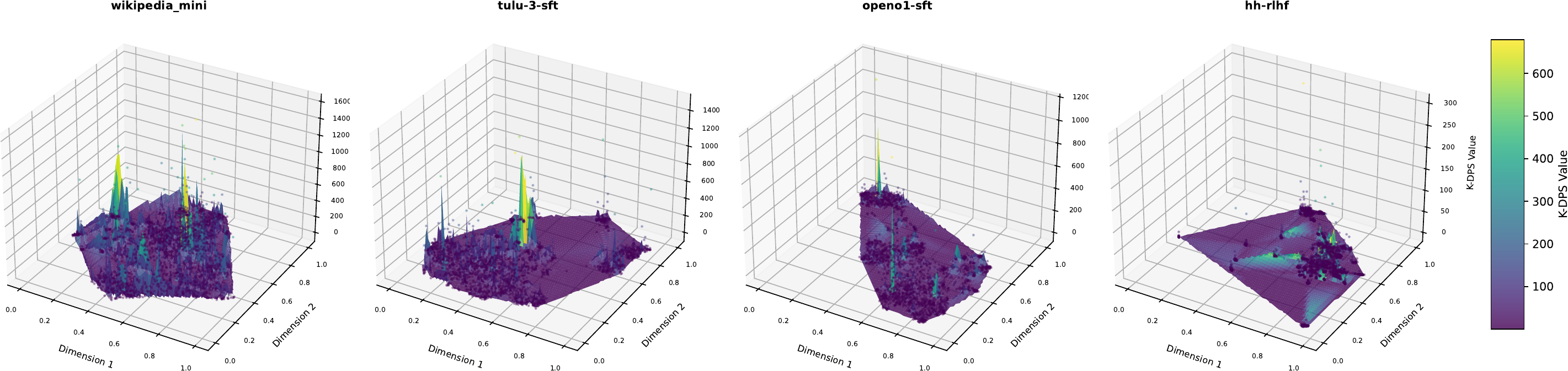}
\caption{Three-dimensional visualization of the $K$-DPS ($K=2,500$)
  for Llama-3.2-1B on four datasets.}
\label{fig:vis-3d}
\end{figure*}

\begin{figure*}[h]
\centering
\includegraphics[width=0.90\linewidth]{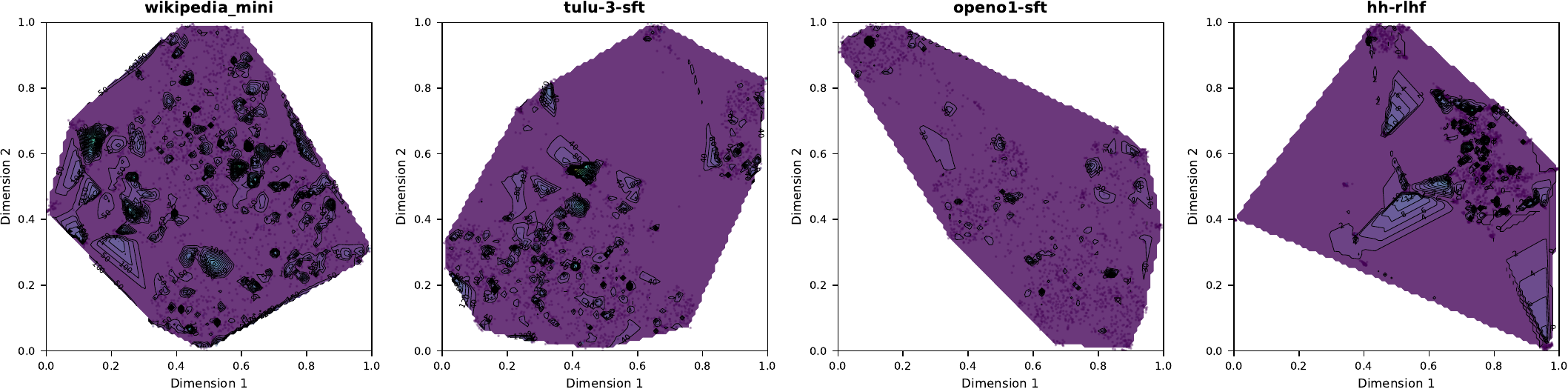}
\caption{Contour visualization of 2,500-grained decision potential
  surface for Llama-3.2-1B on four datasets with linear interpolation.}
\label{fig:vis-contour-linear}
\end{figure*}

\begin{figure*}[h]
\centering
\includegraphics[width=0.90\linewidth]{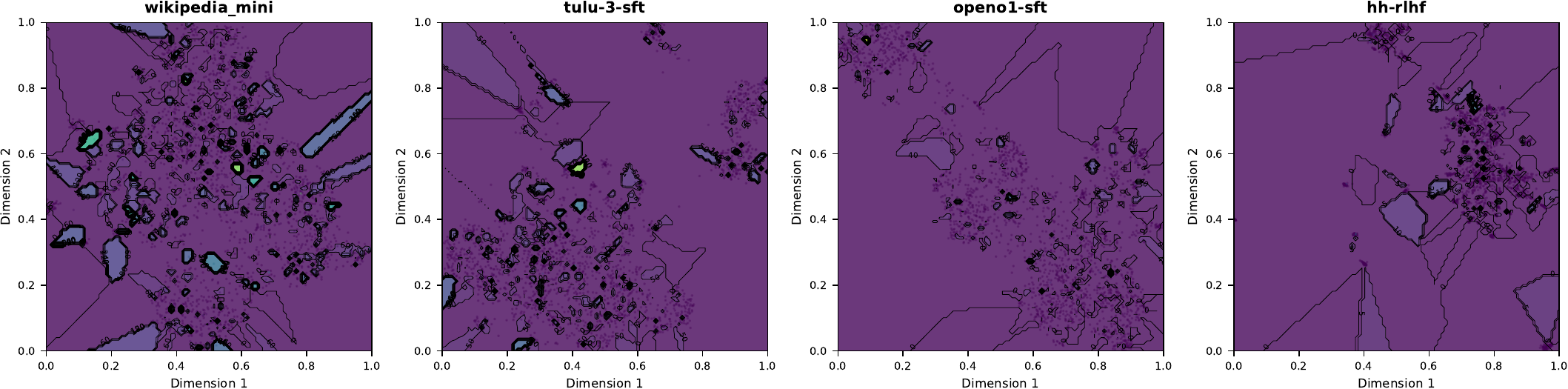}
\caption{Contour visualization of 2,500-grained decision potential
  surface of Llama-3.2-1B on four datasets with nearest interpolation.}
\label{fig:vis-contour-nearest}
\end{figure*}

\begin{figure*}[h]
\centering
\includegraphics[width=0.90\linewidth]{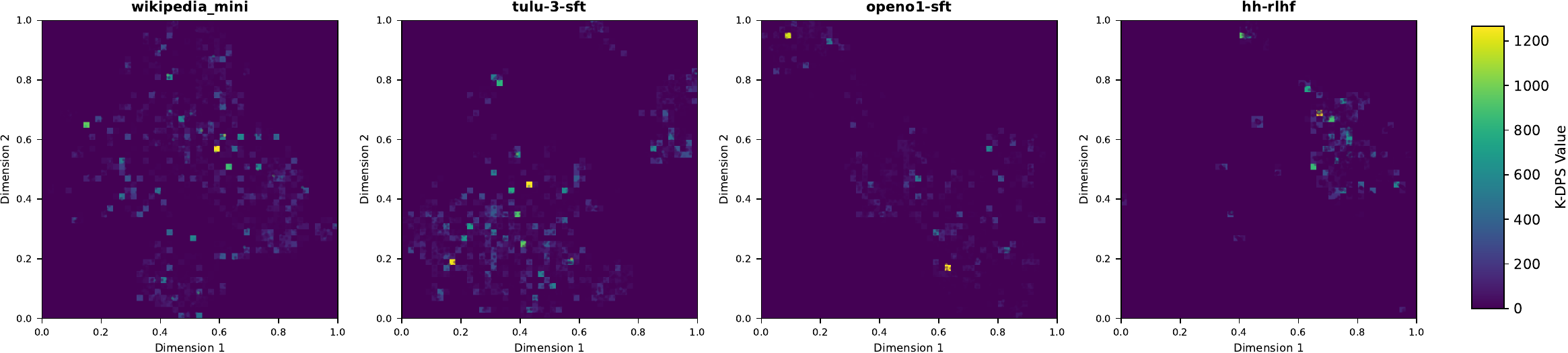}
\caption{Heatmap visualization of the decision potential surface on
  four datasets.}
\label{fig:vis-contour-heatmap}
\end{figure*}


\begin{figure*}[h]
\centering
\includegraphics[width=0.90\linewidth]{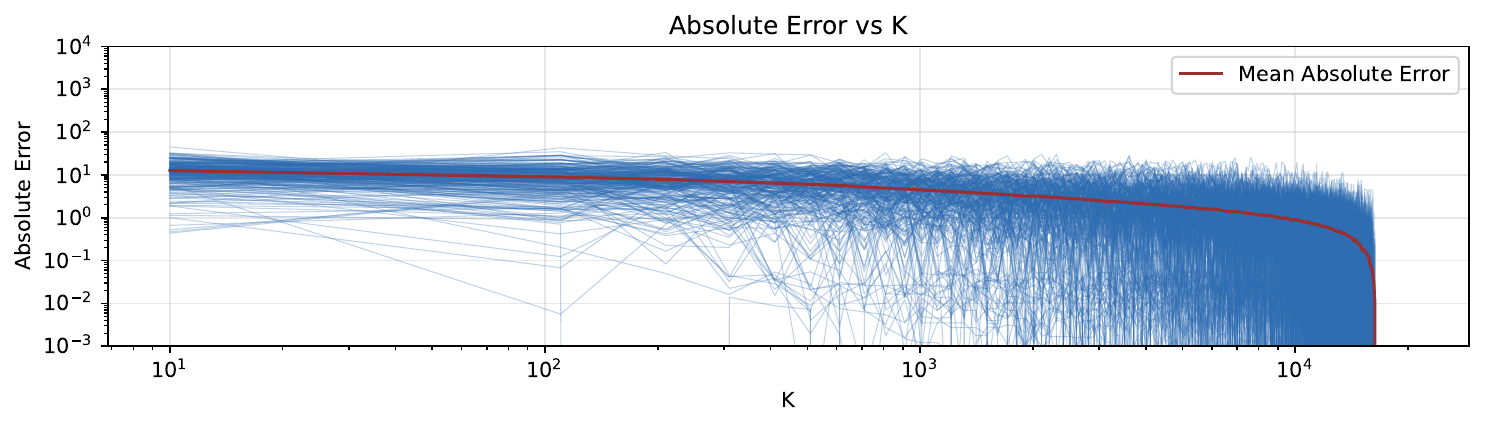}
\caption{Effect of sampling size $K$ on the absolute error between the
  reference $K$-DPS (computed with $K=20,000$) and $K$-DPS values for
  varying $K$. Each blue
line represents a trend of absolute error across input samples.}
\label{fig:varyk-error-log}
\end{figure*}

\section{Proofs}
\input{proof-mcls}
\input{proof-dps}
\input{proof-error-bound}

\section{Discussions}

\subsection{Quantitative Analysis of Top-2 Sequence Similarity}

As discussed in Section~\ref{sec:deg}, the top two completions sampled
under $K$-DPS can occasionally be nearly identical. We provide a
quantitative assessment here. For all candidate sequences used to
construct Figure~\ref{fig:vis-contour-linear}, we measured the
normalized Levenshtein edit distance between top-1 and top-2
completions.

\begin{table}[htbp]
\centering
\caption{Average normalized edit distance between top-1 and top-2
  completions as a function of $K$-DPS confidence score range.}
\begin{tabular}{l|c}
\Xhline{1.25pt}
\textbf{$K$-DPS Score Range} & \textbf{Avg. Normalized Edit Distance} \\
\hline
$<0.1$  & 0.15 \\
$<0.5$  & 0.20  \\
$<1.0$  & 0.23 \\
$<5.0$  & 0.30 \\ 
$<10.0$  & 0.32 \\ 
\Xhline{1.25pt}
\end{tabular}
\label{tab:edit-dis}
\end{table}

As shown in Table~\ref{tab:edit-dis}, even in the highest-confidence
regime ($K$-DPS $<$ 0.1), the top-2 sequences differ by roughly 15\%
of tokens on average; near decision boundaries (higher $K$-DPS),
divergence reaches 30--32\%, confirming that the top two candidates
are typically far from trivial variants.

\subsection{Effect of Filtering on the Local Candidate Set Bound}
\label{sec:filter-bound}

We next clarify whether token- or sequence-level filtering changes the
theoretical guarantees. The answer depends on how filtering is used.
If filtering is only a diagnostic or visualization post-processing
step after computing the raw $K$-DPS, then none of the raw DPS
theorems are changed. The quantities in
Theorems~\ref{th:error}--\ref{th:tail-error} and
Theorem~\ref{th:local-candidate-bound} are still computed from the
unfiltered language-model distribution $P_f(\cdot|\mathbf{x})$.

If filtering is used inside candidate selection, it should be viewed as
defining a filtered or coarsened DPS variant. Let
$a_{\mathbf{x}}(\mathbf{y})\in\{0,1\}$ denote an acceptance rule
for prompt $\mathbf{x}$, such as removing near-duplicate completions
according to edit distance or a semantic-similarity threshold. The
filtered candidate set is
\begin{equation}
\mathcal{Y}_K^{a}=\{\mathbf{y}\in\mathcal{Y}_K:
a_{\mathbf{x}}(\mathbf{y})=1\}.
\end{equation}
Assume that the top sequence is retained,
$a_{\mathbf{x}}(\mathbf{y}_{1*})=1$, and define the accepted local
near-second set
\begin{equation}
\begin{aligned}
\mathcal{Y}_{2,\eta}^{a}(\mathbf{x})=\{&
\mathbf{y}\in\mathcal{V}^{N_r}: a_{\mathbf{x}}(\mathbf{y})=1,\,
\mathbf{y}\neq\mathbf{y}_{1*},\\
&\log P_f(\mathbf{y}|\mathbf{x})
\geq \log P_f(\mathbf{y}_{2*}|\mathbf{x})-\eta\}.
\end{aligned}
\end{equation}
Let
\begin{equation}
p_{2,\eta}^{a}(\mathbf{x})=
\sum_{\mathbf{y}\in\mathcal{Y}_{2,\eta}^{a}(\mathbf{x})}
P_f(\mathbf{y}|\mathbf{x})
\end{equation}
be the probability mass of accepted candidates that are still within
$\eta$ nats of the raw second-best sequence.

The proof of Theorem~\ref{th:local-candidate-bound} then applies
without modification after replacing
$p_{2,\eta}(\mathbf{x})$ by $p_{2,\eta}^{a}(\mathbf{x})$. Specifically,
on the event
\begin{equation}
E_\eta^{a}=
\{\mathbf{y}_{1*}\in\mathcal{Y}_K^{a}\}
\cap
\{\mathcal{Y}_K^{a}\cap
\mathcal{Y}_{2,\eta}^{a}(\mathbf{x})\neq\emptyset\},
\end{equation}
the filtered sample still contains the raw top sequence and at least
one accepted sequence whose log-likelihood is within $\eta$ of
$\log P_f(\mathbf{y}_{2*}|\mathbf{x})$. Hence
\begin{equation}
|\Phi_f^{K,a}(\mathbf{x})-\Phi_f^\infty(\mathbf{x})|
\leq \eta(2\Delta_\infty(\mathbf{x})+\eta)
\end{equation}
with probability at least
\begin{equation}
1-\left(1-P_f(\mathbf{y}_{1*}|\mathbf{x})\right)^K
-\left(1-p_{2,\eta}^{a}(\mathbf{x})\right)^K .
\end{equation}
Thus, \textbf{filtering does not introduce any dependence on the global
diameter $R_K(\mathbf{x})$. It only changes the candidate-discovery
probability through the accepted local mass
$p_{2,\eta}^{a}(\mathbf{x})$}. If the filter removes many near-second
candidates, then $p_{2,\eta}^{a}(\mathbf{x})$ decreases and a larger
$K$ may be needed; if the filter mainly removes low-probability
duplicates or outliers far from the local top-2 band, the local bound
is essentially unchanged.

There is one important distinction. If the filter deliberately removes
the raw second-best sequence and all accepted candidates within
$\eta$ of it, then the estimator no longer targets the raw sequence-
level DPS; it targets a coarsened semantic DPS. Let
$\mathbf{y}_{2*}^{a}$ be the best accepted non-top sequence and
\begin{equation}
\Delta_\infty^{a}(\mathbf{x})=
\log P_f(\mathbf{y}_{1*}|\mathbf{x})
-
\log P_f(\mathbf{y}_{2*}^{a}|\mathbf{x}).
\end{equation}
The same local candidate-set proof gives
\begin{equation}
|\Phi_f^{K,a}(\mathbf{x})-\Phi_f^{\infty,a}(\mathbf{x})|
\leq \eta(2\Delta_\infty^{a}(\mathbf{x})+\eta),
\end{equation}
where $\Phi_f^{\infty,a}(\mathbf{x})=(\Delta_\infty^{a}(\mathbf{x}))^2$.
Relative to the raw DPS, the only additional term is the intentional
coarsening bias
\begin{equation}
|\Phi_f^{\infty,a}(\mathbf{x})-\Phi_f^\infty(\mathbf{x})|
=
|\Delta_\infty^{a}(\mathbf{x})^2-\Delta_\infty(\mathbf{x})^2|.
\end{equation}
This term is not a sampling failure and is not controlled by
$R_K(\mathbf{x})$; it quantifies the deliberate change of target from
raw sequence-level boundaries to filtered semantic boundaries.
Consequently, filtering does not make the theoretical bound vacuous.
It either leaves the raw local candidate-set bound intact, with
$p_{2,\eta}^{a}$ replacing $p_{2,\eta}$, or defines a separate
coarsened DPS object with the same type of local bound plus an explicit
coarsening bias relative to raw DPS.

\subsection{$K$-DPS versus Model Uncertainty}
We notice that the construction of explicit decision boundaries in
the representation space might exhibit connections with several core
research areas in LLMs, particularly confidence
estimation and uncertainty quantification
(UQ)~\citep{uncer-sur1,uncer-sur2,uncer-sur3,uncer-sur4,uncer-sur5,uncer-black-box}.
These uncertainty quantification approaches typically
include verbalized confidence expressed in natural language \citep{vc},
token-level entropy of the output distribution \citep{t-e}, and
semantic entropy computed over semantically equivalent clusters of
multiple generations \citep{t-e,se}, with the latter achieving
state-of-the-art performance in hallucination detection and selective
generation tasks.

While these methods also measure the certainty and confidence of
model decisions, our $K$-DPS decision-boundary construction differs
from them in several fundamental aspects:

First, classical uncertainty quantification techniques
\cite{vc,t-e,se} are essentially heuristic or sampling-based scores
lacking formal theoretical guarantees, whereas $K$-DPS provides provably
conservative classification boundaries with explicit error bounds. It
achieves a precise and meaningful approximation of the decision boundary.
Second, existing UQ methods operate at the instance level and treat
each generation independently, while $K$-DPS explicitly builds and
reasons over distribution-level decision boundaries, enabling global
geometric understanding of the model's reliable support.
In terms of usage, conventional approaches remain largely oblivious to the
location of samples relative to the empirical data manifold, whereas
$K$-DPS deliberately identifies and penalizes anomalous boundary samples
that fall near or outside the observed support of each semantic class.
These distinctions shift the paradigm from post-hoc uncertainty scoring to principled, boundary-aware certification of LLM generations.

Nevertheless, we acknowledge that $K$-DPS and traditional uncertainty
quantification methods indeed share some core insights.
Both paradigms ultimately aim to identify when an LLM's output is
unreliable, whether due to hallucination, out-of-distribution inputs,
adversarial attacks, or memorization-based spurious responses.
Technically, they all ground their analysis in the same internal
representations of the model: prior UQ approaches directly use raw
logits, token probabilities, or hidden states to compute verbalized
confidence or entropy measures, whereas $K$-DPS leverages the DPF as the
theoretical indicator to perform boundary construction.
Consequently, the decision boundary learned by $K$-DPS can be
interpreted as a geometrically principled extension of uncertainty
signals: samples assigned high semantic entropy or low verbalized
confidence often naturally fall into low-density or boundary regions
detected by $K$-DPS, providing a unified explanatory framework for why
existing UQ methods succeed or fail on specific examples.
In practice, the two families of approaches are highly complementary:
uncertainty scores can serve as lightweight pre-filters, while $K$-DPS
offers stricter, certifiable analysis for LLM inference.

\subsection{Effect of Sampling Temperature on DPS}\label{sec:T}

We formalize the effect of temperature on the DPS. Recall that at each
generation step $t$, the LLM produces a logit vector
$z^{(t)}\in\mathbb{R}^V$ from the hidden state, and the next-token
distribution is obtained via softmax: $P_f(y_t\mid\mathbf{x},
y_{<t}) = \text{softmax}(z^{(t)})_{y_t}$. Let
$P_f^{(T)}(\mathbf{y}\mid\mathbf{x})$ denote the temperature-adjusted
distribution with temperature $T>0$, where the logits are scaled
before softmax:
$P_f^{(T)}(y_t\mid\mathbf{x},y_{<t}) = \text{softmax}(z^{(t)}/T)_{y_t}$.
The standard distribution is recovered at $T=1$.

\begin{proposition}[Monotonicity under Temperature]\label{prop:temp}
For any prompt $\mathbf{x}$ and temperature $T_1<T_2$, the ordering of
output sequences by log-probability is preserved. Consequently, the
zero-height isohypse $\mathcal{D}'_{(0,f)}$ is invariant under
temperature changes, and the DPS undergoes a monotone transformation
$\Phi_f^{(\infty,T_2)}(\mathbf{x})=\frac{1}{T_2^2}\Phi_f^{(\infty,T_1)}(\mathbf{x})$.
\end{proposition}

\begin{proof}
At temperature $T$, the log-probability of the generated token $y_t$ at
step $t$ becomes $\log P_f^{(T)}(y_t\mid\mathbf{x},y_{<t})
= \frac{1}{T}z^{(t)}_{y_t} - \log Z^{(T)}_t$, where $Z^{(T)}_t
= \sum_{v\in\mathcal{V}} \exp(z^{(t)}_v/T)$ is the per-step partition
function. Summing over $t=1,\dots,N_r$ gives the sequence
log-probability. Since $z^{(t)}_{y_t}$ is independent of $T$ and
$1/T>0$ is strictly monotone, the ordering of sequences by
log-probability is unchanged, so the identity of $\mathbf{y}_{1*}$ and
$\mathbf{y}_{2*}$ is preserved. The DPF gap becomes
$\Delta_\infty^{(T)}(\mathbf{x})=\frac{1}{T}
\Delta_\infty^{(1)}(\mathbf{x})$, yielding
$\Phi_f^{(\infty,T)}(\mathbf{x})=\frac{1}{T^2}\Phi_f^{(\infty,1)}(\mathbf{x})$.
\end{proof}

Proposition~\ref{prop:temp} implies that temperature scales the DPS
values by $1/T^2$ while preserving the zero-height isohypse.
Higher temperature compresses the surface vertically, reducing the
dynamic range of DPS values; lower temperature amplifies differences.
In both cases the topological structure of isohypses, including the
decision boundary itself, remains identical.

\subsection{Sampling Strategies and DPS}\label{sec:sampling}

In practice, candidate sequences $\mathcal{Y}_K$ are drawn using
decoding strategies rather than the raw model distribution. We discuss the
compatibility of common strategies with $K$-DPS.

$\bullet$
\textbf{Nucleus (top-$p$) sampling} restricts candidates to the
smallest set whose cumulative probability exceeds $p$. Since this set
is a subset of the support of $P_f$, the error bounds in
Theorems~\ref{th:error}--\ref{th:tail-error} remain valid, with
$\varepsilon_{\text{tail}}$ now depending on the truncated
distribution. For typical $p\in[0.9,0.95]$, the truncation is mild and
the bounds hold with slightly adjusted constants.

$\bullet$
\textbf{Top-$k$ sampling} restricts candidates to the $k$ most likely
tokens at each step. When $k\ll V$, the sampling distribution differs
from $P_f$, and the theoretical bounds require modification to account
for the restricted support. However, we recommend $k$ large enough
(e.g., $k\geq 50$) so that the top few tokens at each step are always
included, which is sufficient for accurate DPS estimation at modest
$K$.

$\bullet$
\textbf{Temperature} affects the sampling distribution as analyzed in
Appendix~\ref{sec:T}. When temperature is used only for candidate
generation while evaluating DPF values on the raw ($T=1$) logits, the
theoretical guarantees apply directly. When temperature is applied to
both generation and evaluation, Proposition~\ref{prop:temp} guarantees
that the decision boundary structure is preserved.

Note that the above comparison is about the DPS between the standard
token sampling and these sampling strategies. If we consider the
sampling strategy itself as part of the model, then the above
analysis is unnecessary and the original derivations still apply.

In our experiments (Section~5) we use nucleus sampling with $p=0.9$
and evaluation on raw logits, which balances sample diversity with
theoretical fidelity.

\section{Additional Empirical Analysis}

We provide supplementary empirical checks for the stability of
$K$-DPS estimation. These results address five practical factors that
could affect the estimator: model scale, the normalization of the
estimation error, the maximum decoding length, the sampling strategy,
and the model family.

\subsection{Model-Size Ablation}

We first test whether the sample budget needed for stable $K$-DPS
estimation changes substantially with model size. We evaluate the
Pythia~\citet{DBLP:conf/icml/BidermanSABOHKP23} family, from 70M to 1.4B parameters, on the AdvBench
harmful-prompt set. For each model, we compute the absolute difference
between the $K$-DPS estimate and a reference estimate computed with
$K=2{,}500$, while varying the sampling budget from $K=500$ to
$K=1{,}250$.

\begin{table}[htbp]
\centering
\caption{Estimation error of $K$-DPS across Pythia model scales.
  Reference $K=2{,}500$. Errors below $10^{-13}$ are at machine precision.}
\small
\begin{tabular}{l|cccc}
\Xhline{1.25pt}
\textbf{Model} & $K{=}500$ & $K{=}750$ & $K{=}1{,}000$ & $K{=}1{,}250$ \\
\hline
70M   & 240.91 & 105.86 & $1.17{\times}10^{-14}$ & $1.17{\times}10^{-14}$ \\
160M  & 212.16 & 90.98  & $1.38{\times}10^{-14}$ & $1.38{\times}10^{-14}$ \\
410M  & 213.60 & 99.94  & $9.33{\times}10^{-15}$ & $9.33{\times}10^{-15}$ \\
1B    & 155.98 & 65.48  & $8.63{\times}10^{-15}$ & $8.63{\times}10^{-15}$ \\
1.4B  & 160.97 & 66.93  & $1.12{\times}10^{-14}$ & $1.12{\times}10^{-14}$ \\
\Xhline{1.25pt}
\end{tabular}
\label{tab:pythia-ablation}
\end{table}

Table~\ref{tab:pythia-ablation} shows that the estimator stabilizes
across all tested model sizes. Once $K\geq 1{,}000$, the discrepancy
from the $K=2{,}500$ reference estimate is below $10^{-13}$, i.e.,
at numerical precision. The smaller-budget columns also show a
consistent trend: larger models tend to have lower error at the same
$K$. This is consistent with the top completions carrying more
concentrated probability mass, which makes the top-2 gap easier to
recover from samples.

\subsection{Relative Error Ratio}

Absolute error alone can be hard to interpret when the reference DPS
values vary in magnitude. We therefore also examine the relative error
ratio
$|\Phi_f^K(\mathbf{x})-\Phi_f^{20,000}(\mathbf{x})| /
(\Phi_f^{20,000}(\mathbf{x})+\epsilon)$, with $\epsilon=10^{-8}$. This
metric measures the estimation error relative to the scale of the
reference value and is therefore complementary to the absolute-error
curves in the main text.

\begin{figure}[htbp]
    \centering
    \includegraphics[width=0.95\linewidth]{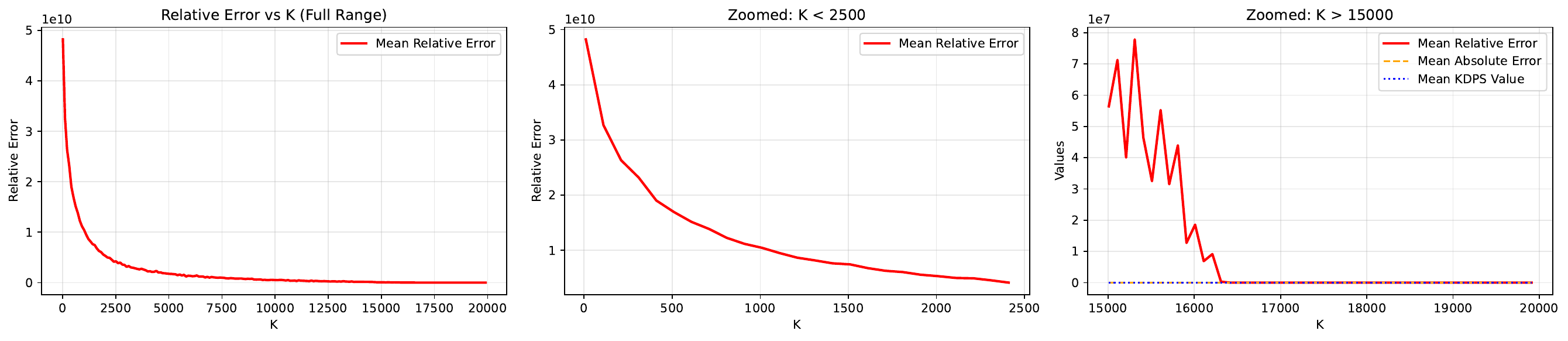}
    \caption{Relative error ratio of $K$-DPS as a function of the
    sampling budget $K$. The reference value is computed with
    $K=20{,}000$.}
    \label{fig:varyk-rel-error}
\end{figure}

Figure~\ref{fig:varyk-rel-error} shows that the relative error
decreases rapidly with $K$ and is already close to zero before
$K=2{,}500$ on the examined inputs. The trend is consistent with the
$1/\sqrt{K}$ convergence rate predicted by Theorem~\ref{th:error}.
Thus the convergence is not an artifact of using an absolute scale:
the estimator is also accurate relative to the magnitude of the DPS
value itself.

\subsection{Long-Generation Tail Probability and Length Budget}

We next clarify the role of generation length in the tail-probability
term. The main issue is not the vocabulary size alone, but the joint
probability of an exact generated sequence. For very long,
high-entropy generation, this joint probability can be small even when the
model assigns high probability to most individual tokens. To make this
point explicit, we evaluated top-1 sequence probabilities under an
8K-token maximum decoding setting on long-form reasoning and generation prompts.

\begin{table}[htbp]
\centering
\caption{Top-1 sequence probability under an 8K-token decoding cap.
The cap is the maximum allowed length; the realized generation length
can be shorter.}
\scriptsize
\begin{tabular}{l|ccc}
\Xhline{1.25pt}
\textbf{Model} & \makecell{\textbf{Avg.}\\\textbf{Len.}} &
\makecell{\textbf{Mean Top-1}\\\textbf{Seq. Prob.}} &
\makecell{\textbf{Avg. Token}\\\textbf{Prob.}} \\
\hline
Llama-3.1-8B & 7505 & $5.86{\times}10^{-13}$ & 0.967 \\
Llama-3.2-3B & 3572 & $1.44{\times}10^{-3}$  & 0.856 \\
Qwen3.5-9B   & 1913 & $8.45{\times}10^{-3}$  & 0.892 \\
Qwen2.5-7B   & 6058 & $7.14{\times}10^{-5}$  & 0.938 \\
\Xhline{1.25pt}
\end{tabular}
\label{tab:long-generation-tail}
\end{table}

Table~\ref{tab:long-generation-tail} supports a more precise reading
of the theory. In long open-ended settings, the sequence-level top-1
probability can indeed be very small: for example, the mean top-1
sequence probability for Llama-3.1-8B is $5.86{\times}10^{-13}$. This
means that the original tail-probability bound may require a much
larger sample budget if it is interpreted as a guarantee for exact
8K-token sequence recovery. This is a limitation of exact
sequence-level boundary construction in high-entropy generation, and
we state it explicitly.

At the same time, \textbf{these numbers should not be interpreted by dividing
the log sequence probability by the maximum cap of 8K tokens in every
case}. The cap is not the realized length: the average realized lengths
in Table~\ref{tab:long-generation-tail} range from 1913 to 7505
tokens. Moreover, the average per-token probabilities remain moderate
to high. The small joint probabilities arise from multiplying many
token probabilities over long outputs, not from every token having an
implausibly tiny probability. This highlights the practical scope of
exact sequence-level DPS: its sampling guarantee is most meaningful
when the relevant top sequences have non-negligible probability mass,
whereas very long high-entropy completions are better analyzed through
localized or prefix-conditioned variants.

For this reason, our empirical $K$-DPS claims are intended for the
bounded-length response regimes studied in the main experiments. In
very long open-ended generation, token-level DPS
(Section~\ref{sec:token-dps}, $N_r=1$), prefix-conditioned DPS, or
shorter-window sequence DPS is the appropriate diagnostic because the
candidate mass is concentrated locally at each prefix rather than over
an entire 8K-token completion.

\paragraph{Bounded-length ablation.}
As a separate sanity check, we also vary a moderate maximum generation
length from 16 to 256 tokens using Llama-3.2-1B on Wikipedia Mini.
Table~\ref{tab:length-ablation} reports the resulting $K$-DPS values
and absolute errors relative to the corresponding high-budget reference
estimator.

\begin{table}[htbp]
\centering
\caption{Effect of maximum generation length on $K$-DPS estimation.
The length column is the decoding cap, not the average realized
generation length.}
\begin{tabular}{c|cc}
\Xhline{1.25pt}
\textbf{Max. Length} & \textbf{$K$-DPS Value} & \textbf{Abs. Error} \\
\hline
16  & 0.002 & $\approx 0$ \\
32  & 0.72  & $\approx 0$ \\
64  & 0.66  & $\approx 0$ \\
128 & 0.29  & $\approx 0$ \\
256 & 0.41  & $\approx 0$ \\
\Xhline{1.25pt}
\end{tabular}
\label{tab:length-ablation}
\end{table}

Within this bounded-length range, increasing the maximum length does
not cause a systematic increase in estimation error. This experiment
therefore supports the stability of $K$-DPS for the response lengths
used in our main empirical studies, but it should not be read as a
claim that the same sample budget resolves arbitrary 8K-token
open-ended generation.

\subsection{Sampling-Strategy Sensitivity}

We finally test whether the estimator depends on the decoding strategy
used to construct the candidate set. Table~\ref{tab:sampling-ablation}
summarizes the comparison.

\begin{table}[htbp]
\centering
\caption{Effect of sampling strategy on $K$-DPS estimation. Greedy
search is degenerate for $K$-DPS because it repeatedly returns a single
candidate and therefore cannot estimate a top-2 sample gap.}
\begin{tabular}{l|cc}
\Xhline{1.25pt}
\textbf{Sampling Strategy} & \textbf{$K$-DPS Value} & \textbf{Abs. Error} \\
\hline
Greedy search & -- & -- \\
Nucleus sampling & 0.144 & $\approx 0$ \\
Top-$k$ clipping & 0.25 & $\approx 0$ \\
\Xhline{1.25pt}
\end{tabular}
\label{tab:sampling-ablation}
\end{table}

Greedy search is degenerate for this purpose: repeated decoding returns
the same highest-probability sequence, so the sample set does not
contain the second candidate needed to estimate a top-2 gap. Stochastic
strategies such as nucleus sampling and top-$k$ clipping avoid this
collapse by preserving candidate diversity. Under the same likelihood
evaluation protocol, both stochastic strategies produce essentially
zero absolute error against the reference estimate, supporting our use
of stochastic candidate generation followed by raw-logit likelihood
evaluation.

\section{Visualization Details}\label{sec:supp-vis}

While this paper primarily focuses on the error analysis of
LLM decision boundary construction, our proposed
$K$-DPS can also be used to intuitively visualize both the decision
boundary and the decision potential surface of an LLM under a given
input distribution, as detailed in Section \ref{sec:impli}. In this
section, we detail the settings and the visualization effectiveness.

\noindent
\textbf{Settings.}\label{sec:vis-set}
For visualization, we construct a low-dimensional
representation of the original input distribution $\mathcal{D}'$, typically in two
dimensions to facilitate human understanding. First, we
extract the last hidden state of an input $\mathbf{x}$ from the LLM as
the original embedding of the input point. Next, we apply UMAP with
100 neighbors and a minimum distance of 0.2 for dimensionality
reduction. Finally, we normalize the reduced embeddings to the range
$[0, 1]$ to construct the decision potential surface visualization. For
interpolation, we evaluate nearest, linear, and cubic interpolation
methods to approximate the $K$-DPS values on a mesh grid.

\noindent
\textbf{Role of Projection.}
We emphasize that dimensionality reduction (UMAP) is used
\emph{only for visualization}; all quantitative conclusions in this
paper are based on the $K$-DPS values computed at \textbf{actual} input data
points \textbf{without} any projection. The interpolation over the mesh grid
visualizes the surface, but the numerical claims about convergence,
error, and boundary structure are drawn directly from the sample-level
DPF values. To further validate that the observed trends are not
artifacts of UMAP, we provide visualizations with three complementary
interpolation methods (nearest, linear, cubic) in
Figures~\ref{fig:vis-contour-cubic}, \ref{fig:vis-contour-linear},
and~\ref{fig:vis-contour-nearest}, respectively. The nearest and linear
interpolations preserve the sign and monotonicity of observed
$K$-DPS values in all regions, and the consistent boundary structures
across all three methods confirm that the qualitative results are
robust to the choice of interpolation. Cubic interpolation may produce
slight negative values in sparse regions (e.g., top-left quadrants of
some panels in Figure~\ref{fig:vis-contour-cubic}), which are
interpolation artifacts from the absence of input samples in those
areas, not errors in the underlying $K$-DPS computation.

%% file: proof-mcls.tex
\subsection{Proof of Theorem \ref{th:mcls}}\label{sec:proof-mcls}
\begin{proof}

\noindent
\textbf{Part I: Proof of Equation \ref{eq:mcls-b}.}

We aim to characterize the decision boundary $\mathcal{B}_M^{(f,
  \mathcal{D})}$ for a neural network $f: \mathbb{R}^d \to
\mathbb{R}^M$ in the multi-class classification setting ($M > 2$)
under an input distribution $\mathcal{D} \subseteq
\mathbb{R}^d$. The network is decomposed as $f = \sigma \circ
f_{\text{cls}} \circ f_r$, where:
\begin{itemize}
    \item $f_r: \mathbb{R}^d \to \mathbb{R}^{d_h}$ maps the input
      $\mathbf{x}$ to a latent representation $h =
      f_r(\mathbf{x})$,
    \item $f_{\text{cls}}: \mathbb{R}^{d_h} \to \mathbb{R}^M$ is a
      linear classification head, $f_{\text{cls}}(h) = W_{\text{cls}}
      h + b_{\text{cls}}$, with $W_{\text{cls}} \in \mathbb{R}^{M
        \times d_h}$, $b_{\text{cls}} \in \mathbb{R}^M$,
    \item $\sigma: \mathbb{R}^M \to \mathbb{R}^M$ is the softmax
      function, $\sigma(z)_i = \frac{e^{z_i}}{\sum_{j=1}^M
        e^{z_j}}$, producing probabilities $P = [p_1, p_2, \dots,
      p_M]$ with $\sum_{i=1}^M p_i = 1$.
\end{itemize}

By Definition \ref{def:d-b}, the decision boundary
$\mathcal{B}_M^{(f, \mathcal{D})}$ is the set of inputs $\mathbf{x}
\in \mathcal{D}$ such that there exist at least two classes $m, n
\in \mathcal{M} = \{1, 2, \dots, M\}$, $m \neq n$, with equal and
maximal probabilities:
\begin{equation}
p_m = p_n \geq \max_{o \in \mathcal{M} \setminus \{m, n\}} p_o.
\end{equation}
Since $\sigma$ is the softmax function, $p_m = \sigma(z)_m =
\frac{e^{z_m}}{\sum_{j=1}^M e^{z_j}}$, the condition $p_m = p_n$
implies:
\begin{equation}
\frac{e^{z_m}}{\sum_{j=1}^M e^{z_j}} = \frac{e^{z_n}}{\sum_{j=1}^M
  e^{z_j}} \implies e^{z_m} = e^{z_n} \implies z_m = z_n.
\end{equation}
The logits are given by $z = W_{\text{cls}} h + b_{\text{cls}}$, so:
\begin{equation}
z_m = w_m h + b_m, \quad z_n = w_n h + b_n,
\end{equation}
where $w_m, w_n$ are the $m$-th and $n$-th rows of $W_{\text{cls}}$,
and $b_m, b_n$ are the corresponding entries of
$b_{\text{cls}}$. Thus, $z_m = z_n$ implies:
\begin{equation}
(w_m - w_n) h + (b_m - b_n) = 0.
\end{equation}
Additionally, for $p_m = p_n$ to be maximal, we require $p_m \geq p_o$
for all $o \neq m, n$, which implies: $\forall o \neq m, n$,
\begin{equation}
  \begin{aligned}
&\frac{e^{z_m}}{\sum_{j=1}^M e^{z_j}} \geq \frac{e^{z_o}}{\sum_{j=1}^M
  e^{z_j}} \implies e^{z_m} \geq e^{z_o} \implies z_m \geq z_o.
  \end{aligned}
\end{equation}
Since $z_m = z_n$, this becomes:
\begin{equation}
z_m = z_n \geq z_o, \quad \forall o \neq m, n.
\end{equation}
In the representation space, this translates to:
\begin{equation}
  \begin{aligned}
&(w_m - w_o) h + (b_m - b_o) \geq 0,\\& (w_n - w_o) h + (b_n - b_o)
\geq 0, \quad \forall o \neq m, n.
  \end{aligned}
\end{equation}
For each pair $m, n \in \mathcal{M}$, $1 \leq m < n \leq M$, define:
\begin{equation}
  \begin{aligned}
&\mathcal{B}_{mn} = \{ h \in \mathbb{R}^{d_h} \mid (w_m - w_n) h +
  (b_m - b_n) = 0, \\& z_m = z_n \geq z_o \, \forall o \neq m, n, \, h
  = f_r(\mathbf{x}), \, \mathbf{x} \in \mathcal{D} \}.
  \end{aligned}
\end{equation}
The decision boundary is the union of all such pairwise boundaries:
\begin{equation}
\mathcal{B}_M^{(f, \mathcal{D})} = \bigcup_{1 \leq m < n \leq M} \mathcal{B}_{mn}.
\end{equation}

\noindent
\textbf{Part II: Voronoi Cells.}

Each $\mathcal{B}_{mn}$ is a $(d_h - 1)$-dimensional hyperplane in
$\mathbb{R}^{d_h}$ defined by $(w_m - w_n) h + (b_m - b_n) = 0$,
restricted to points where $z_m = z_n \geq z_o$. Geometrically, the
classification region for class $i$ is:
\begin{equation}
  \begin{aligned}
&\mathcal{R}_i = \{ h \in \mathbb{R}^{d_h} \mid w_i h + b_i > w_j
  h + b_j, \\& \forall j \neq i, \, h = f_r(\mathbf{x}), \, \mathbf{x}
  \in \mathcal{D} \}.
  \end{aligned}
\end{equation}
These regions are convex polytopes, as they are defined by the
intersection of half-spaces $(w_i - w_j) h + (b_i - b_j) > 0$. The
boundaries between $\mathcal{R}_m$ and $\mathcal{R}_n$ occur where
$(w_m - w_n) h + (b_m - b_n) = 0$ and $z_m = z_n \geq z_o$, forming
$\mathcal{B}_{mn}$. The collection $\{\mathcal{R}_i\}_{i=1}^M$
partitions the representation space, and the hyperplanes
$\mathcal{B}_{mn}$ form the boundaries of a Voronoi-like partition,
where each $\mathcal{R}_i$ is a Voronoi cell corresponding to class
$i$.

This completes the proof.
\end{proof}

\subsection{Proof of Theorem \ref{th:llm-d-b}}\label{sec:appendix-llm-d-b}
\begin{proof}
We aim to characterize the decision boundary \(\mathcal{B}_{llm}^{(f, \mathcal{D}')}\) of an LLM \(f: \mathcal{V}^{N_q} \to \mathcal{V}^{N_r}\) under an input text distribution \(\mathcal{D}' \subseteq \bigcup_{n_q=1}^{N_q} \mathcal{V}^{n_q}\). The LLM generates a sequence \(\mathbf{y} = [y_1, \dots, y_{N_r}] \in \mathcal{V}^{N_r}\), where \(\mathcal{V} = \{1, 2, \dots, V\}\) is the vocabulary, conditioned on a prompt \(\mathbf{x} \in \mathcal{D}'\). The joint probability of generating \(\mathbf{y}\) is:
\[
P_f(\mathbf{y} | \mathbf{x}) = \prod_{t=1}^{N_r} P_f(y_t | \mathbf{x}, y_1, \dots, y_{t-1}),
\]
where \(P_f(y_t | \mathbf{x}, y_1, \dots, y_{t-1})\) is the probability of predicting token \(y_t\) at step \(t\), modeled as a multi-class classification over \(\mathcal{V}\).

Based on \(\mathbf{y}^* = \arg\max_{\mathbf{y} \in
  \mathcal{V}^{N_r}} P_f(\mathbf{y} | \mathbf{x})\) and Definition \ref{def:d-b},
the decision boundary
\(\mathcal{B}_{llm}^{(f, \mathcal{D}')}\) is the set of prompts
\(\mathbf{x} \in \mathcal{D}'\) where at least two distinct sequences
\(\mathbf{y}_v, \mathbf{y}_w \in \mathcal{V}^{N_r}\) have equal and
maximal joint probabilities:
\[
P_f(\mathbf{y}_v | \mathbf{x}) = P_f(\mathbf{y}_w | \mathbf{x}) \geq \max_{\mathbf{y}_u \in \mathcal{V}^{N_r} \setminus \{\mathbf{y}_v, \mathbf{y}_w\}} P_f(\mathbf{y}_u | \mathbf{x}).
\]

For each pair of distinct sequences \(\mathbf{y}_v, \mathbf{y}_w \in \mathcal{V}^{N_r}\), define:
\begin{equation}\small
  \label{eq:111111}
  \begin{aligned}
&\mathcal{B}_{llm, vw} = \{ \mathbf{x} \in \mathcal{D}' \mid\\& P_f(\mathbf{y}_v | \mathbf{x}) = P_f(\mathbf{y}_w | \mathbf{x}) \geq \max_{\mathbf{y}_u \in \mathcal{V}^{N_r} \setminus \{\mathbf{y}_v, \mathbf{y}_w\}} P_f(\mathbf{y}_u | \mathbf{x}) \}.
  \end{aligned}
\end{equation}
The decision boundary is the union over all such pairs:
\[
\mathcal{B}_{llm}^{(f, \mathcal{D}')} = \bigcup_{\mathbf{y}_v \neq \mathbf{y}_w \in \mathcal{V}^{N_r}} \mathcal{B}_{llm, vw}.
\]
To show this, consider the autoregressive process. For a prompt
\(\mathbf{x}\), the probability \(P_f(\mathbf{y} | \mathbf{x})\)
depends on the token probabilities at each step. Obviously, the predicted sequence \(\mathbf{y}^*\) maximizes \(P_f(\mathbf{y} | \mathbf{x})\). The decision boundary occurs when two sequences \(\mathbf{y}_v\) and \(\mathbf{y}_w\) have equal probabilities, and no other sequence has a higher probability. This implies:
\begin{equation}
  \label{eq:222}
  \begin{aligned}
&P_f(\mathbf{y}_v | \mathbf{x}) \\&= \prod_{t=1}^{N_r} P_f(y_{v,t} | \mathbf{x}, y_{v,1}, \dots, y_{v,t-1}) \\&= \prod_{t=1}^{N_r} P_f(y_{w,t} | \mathbf{x}, y_{w,1}, \dots, y_{w,t-1}) \\&= P_f(\mathbf{y}_w | \mathbf{x}),
  \end{aligned}
\end{equation}
and for all \(\mathbf{y}_u \neq \mathbf{y}_v, \mathbf{y}_w\):
\[
P_f(\mathbf{y}_v | \mathbf{x}) \geq P_f(\mathbf{y}_u | \mathbf{x}).
\]
Since each token prediction is a multi-class classification (as in
Theorem \ref{th:mcls}), the boundary for a single token \(y_t\) is
defined by equal probabilities for the top tokens. For the full
sequence, the boundary \(\mathcal{B}_{llm, vw}\) corresponds to
prompts \(\mathbf{x}\) where the joint probabilities align, which may
occur when the log-probabilities differ at some steps but sum to the
same value. The maximality condition ensures that \(\mathbf{y}_v\) and
\(\mathbf{y}_w\) are the top sequences.

This completes the proof.
\end{proof}


%% file: proof-dps.tex
\subsection{Proof of Theorem \ref{th:0-isohypse}}\label{proof-th:0-isohypse}
\begin{proof}
We aim to prove that the decision boundary $\mathcal{B}_{llm}^{(f,
  \mathcal{D}')}$ defined in Theorem \ref{th:llm-d-b} is equivalent
to the 0-isohypse $\mathcal{D}'_{(0,f)}$ on the decision potential
surface $\mathcal{S}^{(f, \mathcal{D}')}$, and that the regions
separated by this boundary correspond exactly to the Voronoi cells in
the token-combined classification definition.

Recall from Theorem \ref{th:llm-d-b} that the decision boundary is
\begin{equation}
\mathcal{B}_{llm}^{(f, \mathcal{D}')} = \bigcup_{\mathbf{y}_v \neq
  \mathbf{y}_w \in \mathcal{V}^{N_r}} \mathcal{B}_{llm, vw},
\end{equation}
where
\begin{equation}\small
  \begin{aligned}
&\mathcal{B}_{llm, vw} \\&= \{ \mathbf{x} \in \mathcal{D}' \mid\\&
  P_f(\mathbf{y}_v | \mathbf{x}) = P_f(\mathbf{y}_w | \mathbf{x}) \geq
  \max_{\mathbf{y}_u \in \mathcal{V}^{N_r} \setminus \{\mathbf{y}_v,
    \mathbf{y}_w\}} P_f(\mathbf{y}_u | \mathbf{x}) \}.
  \end{aligned}
\end{equation}
This boundary consists of prompts
$\mathbf{x}$ where at least two distinct sequences $\mathbf{y}_v$
and $\mathbf{y}_w$ have equal and maximal joint probabilities,
leading to ambiguity in the predicted output sequence.

From Definition \ref{def:ds}, the decision potential function is
\begin{equation}
\Phi_f^{\infty}(\mathbf{x}) = \left( \log P_f(\mathbf{y}_1 |
  \mathbf{x}) - \log P_f(\mathbf{y}_2 | \mathbf{x}) \right)^2,
\end{equation}
where $\mathbf{y}_1, \mathbf{y}_2 \in \mathcal{V}^{N_r}$ are the
sequences with the highest and second-highest log-likelihoods,
respectively. The 0-isohypse is defined as
\begin{equation}
\mathcal{D}'_{(0,f)} = \left\{ \mathbf{x} \in \mathcal{D}' \mid
  \Phi_f^{\infty}(\mathbf{x}) = 0 \right\}.
\end{equation}
By definition, $\Phi_f^{\infty}(\mathbf{x}) = 0$ if and only if
$\log P_f(\mathbf{y}_1 | \mathbf{x}) = \log P_f(\mathbf{y}_2 |
\mathbf{x})$, which implies $P_f(\mathbf{y}_1 | \mathbf{x}) =
P_f(\mathbf{y}_2 | \mathbf{x})$. Since $\mathbf{y}_1$ and
$\mathbf{y}_2$ are the top two sequences by log-likelihood, this
equality ensures that
\begin{equation}
P_f(\mathbf{y}_1 | \mathbf{x}) = P_f(\mathbf{y}_2 | \mathbf{x}) \geq
P_f(\mathbf{y}_u | \mathbf{x}), \quad \forall \mathbf{y}_u \neq
\mathbf{y}_1, \mathbf{y}_2,
\end{equation}
satisfying the maximality condition in Theorem \ref{th:llm-d-b}. Thus,
$\mathbf{x} \in \mathcal{D}'_{(0,f)}$ if and only if $\mathbf{x}
\in \mathcal{B}_{llm}^{(f, \mathcal{D}')}$, establishing the set
equivalence
\begin{equation}
\mathcal{B}_{llm}^{(f, \mathcal{D}')} = \mathcal{D}'_{(0,f)}.
\end{equation}

Geometrically, the regions separated by the 0-isohypse are the
connected components of $\mathcal{D}' \setminus
\mathcal{D}'_{(0,f)}$, where each region corresponds to prompts for
which a unique sequence $\mathbf{y}_i$ has the highest
log-likelihood ($\Phi_f^{\infty}(\mathbf{x}) > 0$). These regions
are exactly the Voronoi cells in the sequence-level classification
framework of Theorem \ref{th:llm-d-b}, as each cell consists of
prompts yielding the same maximal sequence. The
0-isohypse forms the boundaries between these cells, partitioning the
prompt space $\mathcal{D}'$ into regions of unambiguous predictions.

This completes the proof.
\end{proof}

\subsection{Proof of Corollary \ref{cor:epsilon-confidence}}\label{proof-cor:epsilon-confidence}
\begin{proof}
We aim to show that for any $\varepsilon > 0$, the input space
$\mathcal{D}'$ is partitioned into three disjoint strata based on
the value of the decision potential function
$\Phi_f^{\infty}(\mathbf{x})$:
\begin{equation}
\mathcal{D}' = \mathcal{D}'_{(>\varepsilon,f)} \sqcup
\mathcal{D}'_{(<\varepsilon,f)} \sqcup \mathcal{D}'_{(\varepsilon,f)},
\end{equation}
where $\sqcup$ denotes disjoint union.

From Definition \ref{def:epsilon-isohypse}, the $\varepsilon$-isohypse is
\begin{equation}
\mathcal{D}'_{(\varepsilon,f)} = \left\{ \mathbf{x} \in \mathcal{D}' \mid
  \Phi_f^{\infty}(\mathbf{x}) = \varepsilon \right\},
\end{equation}
and the other strata are defined as
\begin{equation}
  \begin{aligned}
&\mathcal{D}'_{(>\varepsilon,f)} = \left\{ \mathbf{x} \in \mathcal{D}'
  \mid \Phi_f^{\infty}(\mathbf{x}) > \varepsilon \right\}, \\& 
\mathcal{D}'_{(<\varepsilon,f)} = \left\{ \mathbf{x} \in \mathcal{D}'
  \mid \Phi_f^{\infty}(\mathbf{x}) < \varepsilon \right\}.
  \end{aligned}
\end{equation}
Since $\Phi_f^{\infty}: \mathcal{D}' \to \mathbb{R}_+$ is a
continuous function (assuming log-likelihoods are continuous in the
prompt space), these sets are disjoint and their union covers
$\mathcal{D}'$.

$\bullet$ \emph{$\varepsilon$-confident regions}: For $\mathbf{x} \in
\mathcal{D}'_{(>\varepsilon,f)}$, $\Phi_f^{\infty}(\mathbf{x}) >
\varepsilon$, so
\begin{equation}
\left| \log P_f(\mathbf{y}_1 | \mathbf{x}) - \log P_f(\mathbf{y}_2 |
  \mathbf{x}) \right| > \sqrt{\varepsilon}.
\end{equation}
Since $\mathbf{y}_1$ has the highest log-likelihood, $\log
P_f(\mathbf{y}_1 | \mathbf{x}) - \log P_f(\mathbf{y}_2 | \mathbf{x}) >
\sqrt{\varepsilon}$, meaning the model predicts $\mathbf{y}_1$ with at
least $\sqrt{\varepsilon}$ nats (natural units of information) of
confidence over the next most likely sequence $\mathbf{y}_2$.

$\bullet$ \emph{$\varepsilon$-uncertain regions}: For $\mathbf{x} \in
\mathcal{D}'_{(<\varepsilon,f)}$, $\Phi_f^{\infty}(\mathbf{x}) <
\varepsilon$, so
\begin{equation}
\left| \log P_f(\mathbf{y}_1 | \mathbf{x}) - \log P_f(\mathbf{y}_2 |
  \mathbf{x}) \right| < \sqrt{\varepsilon}.
\end{equation}
Here, the model has low confidence, with a margin less than
$\sqrt{\varepsilon}$ nats between the top two sequences. As $\varepsilon
\to 0$, $\Phi_f^{\infty}(\mathbf{x}) \to 0$, so
$\mathcal{D}'_{(<\varepsilon,f)}$ converges to the 0-isohypse
$\mathcal{D}'_{(0,f)}$, where the margin is zero.

$\bullet$ \emph{$\varepsilon$-isohypse}: For $\mathbf{x} \in
\mathcal{D}'_{(\varepsilon,f)}$, $\Phi_f^{\infty}(\mathbf{x}) =
\varepsilon$, so the confidence margin is exactly $\sqrt{\varepsilon}$
nats, forming the contour that separates confident and uncertain
regions.

The disjointness of the strata follows from the strict inequalities
and equality defining them, and their union covers $\mathcal{D}'$
since $\Phi_f^{\infty}(\mathbf{x}) \geq 0$ for all $\mathbf{x} \in
\mathcal{D}'$.

This completes the proof.
\end{proof}


%% file: proof-error-bound.tex
\subsection{Proof of Theorem \ref{th:error}}\label{proof-th:error}
\begin{proof}
Let $\mathbf{x} \in \mathcal{D}'$ be a fixed input, and let
$\mathcal{Y}_K = \{\mathbf{y}_1, \mathbf{y}_2, \dots, \mathbf{y}_K\}$
be a set of $K$ \emph{i.i.d.} samples drawn from the language model's
output distribution $P_f(\cdot | \mathbf{x})$. The decision
potential function is:
\begin{equation}
\label{eq:10}
\Phi_f^\infty(\mathbf{x}) = \left( \log P_f(\mathbf{y}_{1*} |
  \mathbf{x}) - \log P_f(\mathbf{y}_{2*} | \mathbf{x}) \right)^2,
\end{equation}
where $\mathbf{y}_{1*}$ and $\mathbf{y}_{2*}$ are the top two generated
texts with the highest log-likelihoods over the entire output
space $\mathcal{V}^{N_r}$, and
\begin{equation}
\label{eq:11}
\Phi_f^K(\mathbf{x}) = \left( \log P_f(\mathbf{y}_{1*}^K | \mathbf{x})
  - \log P_f(\mathbf{y}_{2*}^K | \mathbf{x}) \right)^2,
\end{equation}
where $\mathbf{y}_{1*}^K$ and $\mathbf{y}_{2*}^K$ are the top two
generated texts within $\mathcal{Y}_K$.
We aim to bound the error $|\Phi_f^K(\mathbf{x}) -
\Phi_f^\infty(\mathbf{x})|$ with probability at least $1 - \delta -
2\varepsilon_{\text{tail}}$ for $\delta\in (0,1)$.

\noindent
\textbf{\emph{Step 1}: Preliminary.}
Define:
\begin{equation}
\label{eq:12}
\begin{aligned}
&\Delta_\infty(\mathbf{x}) = \log P_f(\mathbf{y}_{1*} | \mathbf{x}) -
  \log P_f(\mathbf{y}_{2*} | \mathbf{x}),\\
&\Delta_K(\mathbf{x}) = \log P_f(\mathbf{y}_{1*}^K | \mathbf{x}) -
  \log P_f(\mathbf{y}_{2*}^K | \mathbf{x}).\\
\end{aligned}
\end{equation}
Thus, $\Phi_f^\infty(\mathbf{x}) = (\Delta_\infty(\mathbf{x}))^2$ and
$\Phi_f^K(\mathbf{x}) = (\Delta_K(\mathbf{x}))^2$. The error can be
expressed as:
\begin{equation}
\label{eq:13}
\begin{aligned}
&|\Phi_f^K(\mathbf{x}) - \Phi_f^\infty(\mathbf{x})| \\&=
                                                     |(\Delta_K(\mathbf{x}))^2
                                                     -
                                                     (\Delta_\infty(\mathbf{x}))^2|
                                                     \\&=
  |\Delta_K(\mathbf{x}) - \Delta_\infty(\mathbf{x})| \cdot
  |\Delta_K(\mathbf{x}) + \Delta_\infty(\mathbf{x})|.
\end{aligned}
\end{equation}
Since $\mathcal{Y}_K$ is finite, $\mathbf{y}_{1*}^K$ and
$\mathbf{y}_{2*}^K$ are the top-2 outputs in $\mathcal{Y}_K$, which
may not include $\mathbf{y}_{1*}$ or $\mathbf{y}_{2*}$.
Define:
\begin{equation}
\label{eq:14}
R_K(\mathbf{x}) = \log P_f(\mathbf{y}_{1*}^K | \mathbf{x}) -
\min_{\mathbf{y} \in \mathcal{Y}_K} \log P_f(\mathbf{y} | \mathbf{x}),
\end{equation}
which represents the \textit{diameter} of
log-likelihoods in $\mathcal{Y}_K$.
\begin{lemma}[$\Pr (\mathbf{y}_{1*} \notin \mathcal{Y}_{K}) \leq
\varepsilon_{\text{tail}}$]\label{th:eps}
Define the tail probability $\varepsilon_{\text{tail}}$ as:
$\varepsilon_{\text{tail}} = \left(1 - P_f(\mathbf{y}_{1*}^K | \mathbf{x})\right)^K$.
Then, we have $\Pr (\mathbf{y}_{1*} \notin \mathcal{Y}_{K}) \leq
\varepsilon_{\text{tail}}$,
\end{lemma}

\emph{A short proof of Lemma \ref{th:eps}:} As $\mathbf{y}_{k}\in\mathcal{Y}_{K}$
are \emph{i.i.d.}, we know that with $K$ samples of
$\mathbf{y}\sim P_{f}(\cdot|\mathbf{x})$ the probability that we
cannot obtain $\mathbf{y}_{1*}$ obeys a geometric distribution, i.e.,
\begin{equation}
\label{eq:21}
\Pr (\mathbf{y}_{1*}\notin \mathcal{Y}_{K})=(1-P_{f}(\mathbf{y}_{1*}|\mathbf{x}))^{K}.
\end{equation}
As $P_{f}(\mathbf{y}_{1*}|\mathbf{x})\geq P_{f}(\mathbf{y}_{1*}^{K}|\mathbf{x})$, then we have
\begin{equation}
  \label{eq:22}
  \begin{aligned}
&\Pr (\mathbf{y}_{1*} \notin \mathcal{Y}_{K})\\&=
(1-P_{f}(\mathbf{y}_{1*}|\mathbf{x}))^{K} \leq (1-P_{f}(\mathbf{y}_{1*}^{K}|\mathbf{x}))^{K}=
\varepsilon_{\text{tail}},
  \end{aligned}
\end{equation}
which ends the proof.

Based on Lemma \ref{th:eps}, we know that $\varepsilon_{\text{tail}}$
 bounds the probability that the true top output
$\mathbf{y}_{1*}$ is not included in $\mathcal{Y}_K$.

\noindent
\textbf{\emph{Step 2}: Bounding $|\Delta_{K}(\mathbf{x})-\Delta_{\infty}(\mathbf{x})|$.}

Since $\Delta_K(\mathbf{x})$ is computed over a random sample, we consider using
\emph{concentration inequalities} to bound the deviation
$|\Delta_K(\mathbf{x}) - \Delta_\infty(\mathbf{x})|$. The
log-likelihoods $\log P_f(\mathbf{y}_k | \mathbf{x})$ for
$\mathbf{y}_k \in \mathcal{Y}_K$ are \emph{i.i.d.}, and they are
bounded within the diameter $R_K(\mathbf{x})$. By \emph{Hoeffding's
inequality}, the deviation of the sample maximum log-likelihood from
its expected maximum is bounded. Specifically, for the top-1
log-likelihood $\forall~t>0$, we have:
\begin{equation}\small
\label{eq:16}
\Pr\left( \left| \log P_f(\mathbf{y}_{1*}^K | \mathbf{x}) - \log
    P_f(\mathbf{y}_{1*} | \mathbf{x}) \right| > t \right) \leq 2
\exp\left( -\frac{2Kt^2}{R_K^2(\mathbf{x})} \right).
\end{equation}
Similarly, for the second-highest log-likelihood, a similar bound
applies.

Combining these, we have:
\begin{equation}
\label{eq:17}
\begin{aligned}
 &|\Delta_K(\mathbf{x}) - \Delta_\infty(\mathbf{x})|\\&=
|\left(\log P_f(\mathbf{y}_{1*}^K | \mathbf{x}) -\log
                                                      P_f(\mathbf{y}_{2*}^K
                                                      |
                                                      \mathbf{x})\right)-\\&\quad\quad \left(\log
                                                      P_f(\mathbf{y}_{1*}
                                                      | \mathbf{x}) -
                                                      \log
                                                      P_f(\mathbf{y}_{2*}
                                                      |
                                                      \mathbf{x})\right)|\\
&=|\left(\log P_f(\mathbf{y}_{1*}^K | \mathbf{x})-\log
  P_f(\mathbf{y}_{1*} | \mathbf{x})\right)-\\&\quad\quad\left(\log
  P_f(\mathbf{y}_{2*}^K | \mathbf{x})-\log P_f(\mathbf{y}_{2*} |
  \mathbf{x})\right)|.
\end{aligned}
\end{equation}
Based on the triangle inequality $|a-b|\leq |a| + |b|$ when $a,b\in
\mathbb{R}$, we know that
\begin{equation}
\label{eq:delta-triangle}
\begin{aligned}
  &|\Delta_K(\mathbf{x}) - \Delta_\infty(\mathbf{x})|\\&=
|\left(\log P_f(\mathbf{y}_{1*}^K | \mathbf{x})-\log
                                                      P_f(\mathbf{y}_{1*}
                                                      |
                                                      \mathbf{x})\right)-\\&\left(\log
                                                      P_f(\mathbf{y}_{2*}^K
                                                      |
                                                      \mathbf{x})-\log
                                                      P_f(\mathbf{y}_{2*}
                                                      |
                                                      \mathbf{x})\right)|\\
  &\leq
|\log P_f(\mathbf{y}_{1*}^K | \mathbf{x})-\log P_f(\mathbf{y}_{1*} |
    \mathbf{x})|+\\&|\log P_f(\mathbf{y}_{2*}^K | \mathbf{x})-\log
    P_f(\mathbf{y}_{2*} | \mathbf{x})|.
\end{aligned}
\end{equation}
To bound $|\Delta_K(\mathbf{x}) - \Delta_\infty(\mathbf{x})|$, we aim
to find the maximal probability for the event $|\Delta_K(\mathbf{x}) -
\Delta_\infty(\mathbf{x})|<t'$ with $t'>0$.
Without losing generality,
we set $t'=2t$, where the objective can be reformulated as:
\begin{equation}
\label{eq:23}
\begin{aligned}
&\Pr (|\Delta_K(\mathbf{x}) - \Delta_\infty(\mathbf{x})|<t')\\
&=1-\Pr (|\Delta_K(\mathbf{x}) - \Delta_\infty(\mathbf{x})|\geq t'),
\end{aligned}
\end{equation}
where
\begin{equation}
\label{eq:24}
\begin{aligned}
&\Pr (|\Delta_K(\mathbf{x}) - \Delta_\infty(\mathbf{x})|\geq t')\\
&=\Pr (|\log P_f(\mathbf{y}_{1*}^K | \mathbf{x})-\log
  P_f(\mathbf{y}_{1*} | \mathbf{x})|\geq t \\&\quad\mathbf{~or~} |\log
  P_f(\mathbf{y}_{2*}^K | \mathbf{x})-\log
  P_f(\mathbf{y}_{2*} | \mathbf{x})|\geq t))\\
&\leq\Pr (|\log P_f(\mathbf{y}_{1*}^K | \mathbf{x})-\log
  P_f(\mathbf{y}_{1*} | \mathbf{x})|\geq t)+\\&\quad\Pr (|\log
  P_f(\mathbf{y}_{3*}^K | \mathbf{x})-\log P_f(\mathbf{y}_{2*} |
  \mathbf{x})|\geq t)\\
&\leq 2 \exp\left( -\frac{2Kt^2}{R_K^2(\mathbf{x})} \right)+2
  \exp\left( -\frac{2Kt^2}{R_K^2(\mathbf{x})} \right)\\
&= 4 \exp\left( -\frac{2Kt^2}{R_K^2(\mathbf{x})} \right).
\end{aligned}
\end{equation}
So we have
\begin{equation}
\label{eq:25}
\begin{aligned}
&\Pr (|\Delta_K(\mathbf{x}) - \Delta_\infty(\mathbf{x})|<t')\\
&=1-\Pr (|\Delta_K(\mathbf{x}) - \Delta_\infty(\mathbf{x})|\geq t')\\
&\geq 1- 4 \exp\left( -\frac{2Kt^2}{R_K^2(\mathbf{x})} \right).
\end{aligned}
\end{equation}
Suppose we have at least $1-\delta$ probability to support this
event stands, we have
\begin{equation}
\label{eq:26}
\begin{aligned}
&~~~~~~1- 4 \exp\left( -\frac{2Kt^2}{R_K^2(\mathbf{x})} \right)\geq 1-\delta\\
&\Leftrightarrow 4 \exp\left( -\frac{2Kt^2}{R_K^2(\mathbf{x})} \right)\leq \delta\\
&\Leftrightarrow \exp\left( -\frac{2Kt^2}{R_K^2(\mathbf{x})} \right)\leq \frac{\delta}{4}\\
&\Leftrightarrow -\frac{2Kt^2}{R_K^2(\mathbf{x})}\leq \log\frac{\delta}{4}\\
&\Leftrightarrow \frac{2Kt^2}{R_K^2(\mathbf{x})}\geq -\log\frac{\delta}{4}\\
&\Leftrightarrow \frac{2Kt^2}{R_K^2(\mathbf{x})}\geq \log\frac{4}{\delta}\\
&\Leftrightarrow t^{2}\geq \frac{R_K^2(\mathbf{x})}{2K}\log\frac{4}{\delta}\\
&\Leftrightarrow t\geq |{R_K(\mathbf{x})}\sqrt{\frac{\log(4/\delta)}{2K}}|\\
&\Leftrightarrow t\geq {R_K(\mathbf{x})}\sqrt{\frac{\log(4/\delta)}{2K}}.
\end{aligned}
\end{equation}
In other words, $\forall~t>0$ we bound $\Pr (|\Delta_K(\mathbf{x}) -
\Delta_\infty(\mathbf{x})|<t)$ with probability at least $1 - \delta$
when:
\begin{equation}
\label{eq:18}
t = R_K(\mathbf{x}) \sqrt{\frac{\log(4/\delta)}{2K}}.
\end{equation}
\noindent
\textbf{\emph{Step 3}: Bounding
  $|\Delta_{K}(\mathbf{x})+\Delta_{\infty}(\mathbf{x})|$.}
\begin{assumption}[Bounded Population Gap]\label{th:M-RK}
There exists a constant $M > 0$ such that for any $\mathbf{x}$, the population top-2 gap satisfies:
\begin{equation}
\Delta_\infty(\mathbf{x}) = \log P_f(\mathbf{y}_{1*}|\mathbf{x}) - \log P_f(\mathbf{y}_{2*}|
 \mathbf{x}) \leq M
\end{equation}
Then we assume that
\begin{equation}
\label{eq:27}
M \leq R_{K}(\mathbf{x})
\end{equation}
when $K\gg 1$.
\end{assumption}
This assumption is reasonable as most practical language models do not
have extremely large differences between top-2 probabilities, and the
probability differences between top-2 would be much smaller than the
range of between the top-1 and the sample with the minimal probability 
in the sampling set.
Now we can obtain that:
\begin{equation}
|\Delta_K(\mathbf{x}) + \Delta_\infty(\mathbf{x})| \leq
|\Delta_K(\mathbf{x})| + |\Delta_\infty(\mathbf{x})|
 \leq 2\cdot R_K(\mathbf{x}).
\end{equation}
\medskip
\noindent
\textbf{\emph{Step 4}: Final bound.}

Define the events
\begin{equation}
  \begin{aligned}
&A=\{\mathbf{y}_{1*}\in\mathcal Y_K\text{ and }\mathbf{y}_{2*}\in\mathcal Y_K\},
    \\&
B=\{\mathbf{y}_{1*}\notin\mathcal Y_K\text{ or }\mathbf{y}_{2*}\notin\mathcal Y_K\}.
  \end{aligned}
\end{equation}
Lemma~\ref{th:eps} and a union bound give
\begin{equation}
\Pr(B)\le 2\varepsilon_{\text{tail}}.
\end{equation}
\begin{itemize}
\item On event $A$ we have
$\mathbf{y}_{1*}^K=\mathbf{y}_{1*}$ and
$\mathbf{y}_{2*}^K=\mathbf{y}_{2*}$, hence
\begin{equation}
\Phi_f^K(\mathbf{x})=\Phi_f^\infty(\mathbf{x})
\quad\Longrightarrow\quad
|\Phi_f^K(\mathbf{x})-\Phi_f^\infty(\mathbf{x})|=0.
\end{equation}
\item On event $B$ we use the worst-case gap
\begin{equation}
  \begin{aligned}
&|\Phi_f^K(\mathbf{x})-\Phi_f^\infty(\mathbf{x})|\\
&\leq
|\Delta_K(\mathbf{x}) - \Delta_\infty(\mathbf{x})| \cdot
|\Delta_K(\mathbf{x}) + \Delta_\infty(\mathbf{x})|\\
&\leq
R_K(\mathbf{x}) \sqrt{\frac{\log(4/\delta)}{2K}}\cdot 2R_{K}(\mathbf{x})\\
&=
2R_K^{2}(\mathbf{x}) \sqrt{\frac{\log(4/\delta)}{2K}}.
  \end{aligned}
\end{equation}
\end{itemize}
This completes the proof.
\end{proof}

\subsection{Proof of Theorem \ref{th:local-candidate-bound}}\label{proof-th:local-candidate-bound}
\begin{proof}
Fix an input $\mathbf{x}\in\mathcal{D}'$. Recall that
\begin{equation}
\Delta_\infty(\mathbf{x})=
\log P_f(\mathbf{y}_{1*}|\mathbf{x})
-
\log P_f(\mathbf{y}_{2*}|\mathbf{x})
\end{equation}
and $\Phi_f^\infty(\mathbf{x})=\Delta_\infty^2(\mathbf{x})$.

For $\eta\geq 0$, define
\begin{equation}
\begin{aligned}
\mathcal{Y}_{2,\eta}(\mathbf{x})=\{&
\mathbf{y}\in\mathcal{V}^{N_r}: \mathbf{y}\neq\mathbf{y}_{1*},\\
&\log P_f(\mathbf{y}|\mathbf{x})
\geq \log P_f(\mathbf{y}_{2*}|\mathbf{x})-\eta\}.
\end{aligned}
\end{equation}
and define
\begin{equation}
p_{2,\eta}(\mathbf{x})=
\sum_{\mathbf{y}\in\mathcal{Y}_{2,\eta}(\mathbf{x})}
P_f(\mathbf{y}|\mathbf{x}).
\end{equation}
Consider the candidate-discovery event
\begin{equation}
E_\eta=
\{\mathbf{y}_{1*}\in\mathcal{Y}_K\}
\cap
\{\mathcal{Y}_K\cap\mathcal{Y}_{2,\eta}(\mathbf{x})\neq\emptyset\}.
\end{equation}
On this event, the sample contains a population top-1 sequence, so
the maximal sampled log-likelihood equals the population maximal
log-likelihood:
\begin{equation}
\log P_f(\mathbf{y}_{1*}^K|\mathbf{x})
=
\log P_f(\mathbf{y}_{1*}|\mathbf{x}).
\end{equation}
This statement is about the log-likelihood value; if several
sequences tie for the top probability, $\mathbf{y}_{1*}^K$ may be any
maximizer in $\mathcal{Y}_K$.

Moreover, because $\mathcal{Y}_K$ contains at least one candidate from
$\mathcal{Y}_{2,\eta}(\mathbf{x})$, the sampled second-best candidate
has log-likelihood at least
$\log P_f(\mathbf{y}_{2*}|\mathbf{x})-\eta$:
\begin{equation}
\log P_f(\mathbf{y}_{2*}^K|\mathbf{x})
\geq
\log P_f(\mathbf{y}_{2*}|\mathbf{x})-\eta.
\end{equation}
At the same time, $\mathbf{y}_{2*}$ is the second element in the
population ordering by log-likelihood. Therefore the second-largest
log-likelihood in any sampled subset cannot exceed the population
second-largest log-likelihood:
\begin{equation}
\log P_f(\mathbf{y}_{2*}^K|\mathbf{x})
\leq
\log P_f(\mathbf{y}_{2*}|\mathbf{x}).
\end{equation}
Combining the two inequalities gives
\begin{equation}
\begin{aligned}
0
&\leq
\log P_f(\mathbf{y}_{2*}|\mathbf{x})
-
\log P_f(\mathbf{y}_{2*}^K|\mathbf{x})
\leq \eta .
\end{aligned}
\end{equation}
Therefore
\begin{equation}
\begin{aligned}
\Delta_K(\mathbf{x})-\Delta_\infty(\mathbf{x})
&=
\left(\log P_f(\mathbf{y}_{1*}^K|\mathbf{x})
      -\log P_f(\mathbf{y}_{2*}^K|\mathbf{x})\right)\\
&\quad -
\left(\log P_f(\mathbf{y}_{1*}|\mathbf{x})
      -\log P_f(\mathbf{y}_{2*}|\mathbf{x})\right)\\
&=
\log P_f(\mathbf{y}_{2*}|\mathbf{x})
-
\log P_f(\mathbf{y}_{2*}^K|\mathbf{x}),
\end{aligned}
\end{equation}
and hence
\begin{equation}
0\leq\Delta_K(\mathbf{x})-\Delta_\infty(\mathbf{x})\leq\eta.
\end{equation}
The squared-potential error on $E_\eta$ is then
\begin{equation}
\begin{aligned}
&|\Phi_f^K(\mathbf{x})-\Phi_f^\infty(\mathbf{x})|\\
&=|\Delta_K^2(\mathbf{x})-\Delta_\infty^2(\mathbf{x})|\\
&=(\Delta_K(\mathbf{x})-\Delta_\infty(\mathbf{x}))
  (\Delta_K(\mathbf{x})+\Delta_\infty(\mathbf{x}))\\
&\leq
\eta\left((\Delta_\infty(\mathbf{x})+\eta)
          +\Delta_\infty(\mathbf{x})\right)\\
&=\eta\left(2\Delta_\infty(\mathbf{x})+\eta\right).
\end{aligned}
\end{equation}

It remains to lower-bound the probability of $E_\eta$. By a union
bound,
\begin{equation}
\Pr(E_\eta^c)
\leq
\Pr(\mathbf{y}_{1*}\notin\mathcal{Y}_K)
+
\Pr(\mathcal{Y}_K\cap\mathcal{Y}_{2,\eta}(\mathbf{x})=\emptyset).
\end{equation}
Since the $K$ samples are independent,
\begin{equation}
\Pr(\mathbf{y}_{1*}\notin\mathcal{Y}_K)
=\left(1-P_f(\mathbf{y}_{1*}|\mathbf{x})\right)^K
\end{equation}
and
\begin{equation}
\Pr(\mathcal{Y}_K\cap\mathcal{Y}_{2,\eta}(\mathbf{x})=\emptyset)
=\left(1-p_{2,\eta}(\mathbf{x})\right)^K.
\end{equation}
Thus,
\begin{equation}
\Pr(E_\eta)\geq
1-\left(1-P_f(\mathbf{y}_{1*}|\mathbf{x})\right)^K
-\left(1-p_{2,\eta}(\mathbf{x})\right)^K,
\end{equation}
which proves Equations~\eqref{eq:local-candidate-bound}
and~\eqref{eq:local-candidate-prob}.

Finally, suppose $p_{2,\eta}(\mathbf{x})>0$. Using
$(1-u)^K\leq \exp(-Ku)$ for $u\in[0,1]$, the two failure terms are
each at most $\delta/2$ whenever
\begin{equation}
K\geq \max\left\{
\frac{\log(2/\delta)}{P_f(\mathbf{y}_{1*}|\mathbf{x})},
\frac{\log(2/\delta)}{p_{2,\eta}(\mathbf{x})}
\right\}.
\end{equation}
Under this sufficient condition, $\Pr(E_\eta)\geq 1-\delta$.
\end{proof}

\subsection{Proof of Theorem \ref{th:e-error}}\label{proof-th:e-error}
\begin{proof}
We aim to bound the expected error $\mathbb{E}[|\Phi_f^K(\mathbf{x})
- \Phi_f^\infty(\mathbf{x})|]$ for a fixed input $\mathbf{x} \in
\mathcal{D}'$ and a set $\mathcal{Y}_K = \{\mathbf{y}_1,
\mathbf{y}_2, \dots, \mathbf{y}_K\}$ of $K$ \emph{i.i.d.} samples drawn
from the language model's output distribution $P_f(\cdot |
\mathbf{x})$. Recall that:
\begin{equation}
\label{eq:35}
\begin{aligned}
&\Phi_f^\infty(\mathbf{x}) = \left( \log P_f(\mathbf{y}_{1*} |
  \mathbf{x}) - \log P_f(\mathbf{y}_{2*} | \mathbf{x}) \right)^2,
\\& \Phi_f^K(\mathbf{x}) = \left( \log P_f(\mathbf{y}_{1*}^K |
  \mathbf{x}) - \log P_f(\mathbf{y}_{2*}^K | \mathbf{x}) \right)^2,
\end{aligned}
\end{equation}
where \(\mathbf{y}_{1*}, \mathbf{y}_{2*}\) are the top-2 outputs over
the entire output space \(\mathcal{V}^{N_r}\), and
\(\mathbf{y}_{1*}^K, \mathbf{y}_{2*}^K\) are the top-2 outputs in
\(\mathcal{Y}_K\). Define:
\begin{equation}
  \begin{aligned}
&\Delta_\infty(\mathbf{x}) = \log P_f(\mathbf{y}_{1*} | \mathbf{x}) -
\log P_f(\mathbf{y}_{2*} | \mathbf{x}), \\& \Delta_K(\mathbf{x}) =
\log P_f(\mathbf{y}_{1*}^K | \mathbf{x}) - \log P_f(\mathbf{y}_{2*}^K
| \mathbf{x}),
  \end{aligned}
\end{equation}
so that $\Phi_f^\infty(\mathbf{x}) = (\Delta_\infty(\mathbf{x}))^2$,
$\Phi_f^K(\mathbf{x}) = (\Delta_K(\mathbf{x}))^2$, and the error is:
\begin{equation}
  \begin{aligned}
&|\Phi_f^K(\mathbf{x}) - \Phi_f^\infty(\mathbf{x})| \\&=
|(\Delta_K(\mathbf{x}))^2 - (\Delta_\infty(\mathbf{x}))^2| \\&=
|\Delta_K(\mathbf{x}) - \Delta_\infty(\mathbf{x})| \cdot
|\Delta_K(\mathbf{x}) + \Delta_\infty(\mathbf{x})|.
  \end{aligned}
\end{equation}
By Assumption \ref{th:M-RK}, $|\Delta_\infty(\mathbf{x})| \leq
R_K(\mathbf{x})$, where $R_K(\mathbf{x}) = \log
P_f(\mathbf{y}_{1*}^K | \mathbf{x}) - \min_{\mathbf{y} \in
  \mathcal{Y}_K} \log P_f(\mathbf{y} | \mathbf{x})$ is the
log-likelihood diameter of $\mathcal{Y}_K$. Also,
$|\Delta_K(\mathbf{x})| \leq R_K(\mathbf{x})$, so:
\begin{equation}
|\Delta_K(\mathbf{x}) + \Delta_\infty(\mathbf{x})| \leq
|\Delta_K(\mathbf{x})| + |\Delta_\infty(\mathbf{x})| \leq 2
R_K(\mathbf{x}).
\end{equation}
Thus, the error is bounded by:
\begin{equation}
|\Phi_f^K(\mathbf{x}) - \Phi_f^\infty(\mathbf{x})| \leq
|\Delta_K(\mathbf{x}) - \Delta_\infty(\mathbf{x})| \cdot 2
R_K(\mathbf{x}).
\end{equation}
We compute the expectation:
\begin{equation}
\mathbb{E}[|\Phi_f^K(\mathbf{x}) - \Phi_f^\infty(\mathbf{x})|] \leq 2
R_K(\mathbf{x}) \cdot \mathbb{E}[|\Delta_K(\mathbf{x}) -
\Delta_\infty(\mathbf{x})|].
\end{equation}
Define $Z = |\Delta_K(\mathbf{x}) -
\Delta_\infty(\mathbf{x})|$. From the proof of Theorem \ref{th:error}
(Equation \ref{eq:24}), Hoeffding's inequality gives:
\begin{equation}
\Pr(Z \geq t) \leq 4 \exp\left(-\frac{2Kt^2}{R_K^2(\mathbf{x})}\right).
\end{equation}
The expectation of $Z$ is:
\begin{equation}
\mathbb{E}[Z] = \int_0^\infty \Pr(Z \geq t) \, dt \leq \int_0^\infty 4
\exp\left(-\frac{2Kt^2}{R_K^2(\mathbf{x})}\right) \, dt.
\end{equation}
Substitute $u = \frac{2Kt^2}{R_K^2(\mathbf{x})}$, so $t =
R_K(\mathbf{x}) \sqrt{\frac{u}{2K}}$, $dt =
\frac{R_K(\mathbf{x})}{2\sqrt{2K} \sqrt{u}} \, du$. Then:
\begin{equation}\small
\mathbb{E}[Z] \leq \int_0^\infty 4 e^{-u} \cdot
\frac{R_K(\mathbf{x})}{2\sqrt{2K} \sqrt{u}} \, du = \frac{2
  R_K(\mathbf{x})}{\sqrt{2K}} \int_0^\infty \frac{e^{-u}}{\sqrt{u}} \,
du.
\end{equation}
Since $\int_0^\infty \frac{e^{-u}}{\sqrt{u}} \, du =
\Gamma\left(\frac{1}{2}\right) = \sqrt{\pi}$, we have:
\begin{equation}
\mathbb{E}[Z] \leq \frac{2 R_K(\mathbf{x})}{\sqrt{2K}} \cdot
\sqrt{\pi} = R_K(\mathbf{x}) \sqrt{\frac{2\pi}{K}}.
\end{equation}
Thus:
\begin{equation}
  \begin{aligned}
&\mathbb{E}[|\Phi_f^K(\mathbf{x}) - \Phi_f^\infty(\mathbf{x})|] \\&\leq 2
R_K(\mathbf{x}) \cdot R_K(\mathbf{x}) \sqrt{\frac{2\pi}{K}} \\&= 2
R_K^2(\mathbf{x}) \sqrt{\frac{2\pi}{K}}.
  \end{aligned}
\end{equation}
To account for event $B = \{\mathbf{y}_{1*} \notin \mathcal{Y}_K
\text{ or } \mathbf{y}_{2*} \notin \mathcal{Y}_K\}$ with $\Pr(B)
\leq 2\varepsilon_{\text{tail}}$ (from Lemma \ref{th:eps} and union
bound), we note that on event $A = \{\mathbf{y}_{1*} \in
\mathcal{Y}_K \text{ and } \mathbf{y}_{2*} \in \mathcal{Y}_K\}$, the
error is zero. Thus, we add a conservative term for event $B$, where
the error is at most $2 R_K^2(\mathbf{x})$ (since $|\Delta_K|,
|\Delta_\infty| \leq R_K(\mathbf{x})$, so $|\Phi_f^K(\mathbf{x}) -
\Phi_f^\infty(\mathbf{x})| \leq 2 R_K^2(\mathbf{x})$):
\begin{equation}
  \begin{aligned}
&\mathbb{E}[|\Phi_f^K(\mathbf{x}) - \Phi_f^\infty(\mathbf{x})| \cdot
\mathds{1}_B] \\&\leq 2 R_K^2(\mathbf{x}) \cdot \Pr(B) \leq 2
R_K^2(\mathbf{x}) \cdot 2 \varepsilon_{\text{tail}} = 4
R_K^2(\mathbf{x}) \varepsilon_{\text{tail}},
  \end{aligned}
\end{equation}
where $\mathds{1}_{B}$ is the indicator function which is $1$ only
when event $B$ occurs.

Combining both terms, the expected error is:
\begin{equation}
\mathbb{E}[|\Phi_f^K(\mathbf{x}) - \Phi_f^\infty(\mathbf{x})|] \leq 2
R_K^2(\mathbf{x}) \sqrt{\frac{2\pi}{K}} + 4 R_K^2(\mathbf{x})
\varepsilon_{\text{tail}},
\end{equation}
where $\varepsilon_{\text{tail}} = \left(1 - P_f(\mathbf{y}_{1*}^K |
  \mathbf{x})\right)^K$. This completes the proof.
\end{proof}

\subsection{Proof of Corollary \ref{th:tail-error}}\label{proof-th:tail-error}
\begin{proof}
We aim to bound the tail probability $\Pr(|\Phi_f^K(\mathbf{x}) -
\Phi_f^\infty(\mathbf{x})| \geq \lambda)$ for $\lambda > 0$. Using
the same notation as in Theorem \ref{th:e-error}, we have:
\begin{equation}
|\Phi_f^K(\mathbf{x}) - \Phi_f^\infty(\mathbf{x})| =
|\Delta_K(\mathbf{x}) - \Delta_\infty(\mathbf{x})| \cdot
|\Delta_K(\mathbf{x}) + \Delta_\infty(\mathbf{x})|.
\end{equation}
Since $|\Delta_K(\mathbf{x}) + \Delta_\infty(\mathbf{x})| \leq 2
R_K(\mathbf{x})$, let $Z = |\Delta_K(\mathbf{x}) -
\Delta_\infty(\mathbf{x})|$, so:
\begin{equation}
|\Phi_f^K(\mathbf{x}) - \Phi_f^\infty(\mathbf{x})| \leq Z \cdot 2
R_K(\mathbf{x}).
\end{equation}
Thus:
\begin{equation}
  \begin{aligned}
&\Pr(|\Phi_f^K(\mathbf{x}) - \Phi_f^\infty(\mathbf{x})| \geq \lambda) \\&=
\Pr\left(Z \cdot 2 R_K(\mathbf{x}) \geq \lambda\right) = \Pr\left(Z
  \geq \frac{\lambda}{2 R_K(\mathbf{x})}\right).
  \end{aligned}
\end{equation}
From the proof of Theorem \ref{th:error} (Equation \ref{eq:24}),
Hoeffding's inequality gives:
\begin{equation}
\Pr\left(Z \geq t\right) \leq 4
\exp\left(-\frac{2Kt^2}{R_K^2(\mathbf{x})}\right).
\end{equation}
Set $t = \frac{\lambda}{2 R_K(\mathbf{x})}$:
\begin{equation}
  \begin{aligned}
&\Pr\left(Z \geq \frac{\lambda}{2 R_K(\mathbf{x})}\right) \\&\leq 4
\exp\left(-\frac{2K \cdot \left(\frac{\lambda}{2
        R_K(\mathbf{x})}\right)^2}{R_K^2(\mathbf{x})}\right) \\&= 4
\exp\left(-\frac{K \lambda^2}{2 R_K^4(\mathbf{x})}\right).
  \end{aligned}
\end{equation}
Define events $A = \{\mathbf{y}_{1*} \in \mathcal{Y}_K \text{ and }
\mathbf{y}_{2*} \in \mathcal{Y}_K\}$ and $B = \{\mathbf{y}_{1*} \notin
\mathcal{Y}_K \text{ or } \mathbf{y}_{2*} \notin \mathcal{Y}_K\}$. On
event $A$, the error is zero, so it does not contribute to the tail
probability. On event $B$, with $\Pr(B) \leq
2\varepsilon_{\text{tail}}$ (from Lemma \ref{th:eps} and union bound),
the tail probability is bounded by:
\begin{equation}
  \begin{aligned}
&\Pr(|\Phi_f^K(\mathbf{x}) - \Phi_f^\infty(\mathbf{x})| \geq \lambda)
\\&\leq \Pr\left(\left\{Z \geq \frac{\lambda}{2 R_K(\mathbf{x})}\right\}
  \cap B\right) + \Pr(A).
  \end{aligned}
\end{equation}
Since $\Pr(A) \geq 1 - 2\varepsilon_{\text{tail}}$ and the error is
zero on $A$, we focus on event $B$:
\begin{equation}
  \begin{aligned}
&\Pr\left(\left\{Z \geq \frac{\lambda}{2 R_K(\mathbf{x})}\right\} \cap
  B\right) \\&\leq \Pr\left(Z \geq \frac{\lambda}{2
    R_K(\mathbf{x})}\right) + \Pr(B) \\&\leq 4 \exp\left(-\frac{K
    \lambda^2}{2 R_K^4(\mathbf{x})}\right) + 2
\varepsilon_{\text{tail}}.
  \end{aligned}
\end{equation}
Thus, the tail probability is:
\begin{equation}
  \begin{aligned}
&\Pr\left(|\Phi_f^K(\mathbf{x}) - \Phi_f^\infty(\mathbf{x})| \geq
  \lambda\right) \\&\leq 4 \exp\left(-\frac{K \lambda^2}{2
    R_K^4(\mathbf{x})}\right) + 2 \varepsilon_{\text{tail}},
  \end{aligned}
\end{equation}
where $\varepsilon_{\text{tail}} = \left(1 - P_f(\mathbf{y}_{1*}^K |
  \mathbf{x})\right)^K$. This completes the proof.
\end{proof}
